\documentclass[11pt]{article}
\usepackage{amssymb}
\usepackage{amsmath,amsfonts,amsthm}
	\usepackage{subcaption}
	\usepackage{multirow}
\usepackage{graphicx}
\usepackage{float}
\usepackage{natbib}
\usepackage{bbm}
\usepackage{natbib}
\usepackage{url}

\usepackage[mathscr]{eucal}
\usepackage{mathrsfs}

\newcommand{\be}{\begin{equation}}
\newcommand{\ee}{\end{equation}}
\newtheorem{lemma}{Lemma}
\newtheorem{theorem}{Theorem}
\newtheorem{corollary}{Corollary}

\newtheorem{proposition}{Proposition}

\newtheorem{assumption}{Assumption}

\newtheorem{remark}{Remark}

\usepackage[singlelinecheck=false]{caption}
\usepackage{enumerate}
\usepackage{amsmath}
\usepackage{natbib}
\bibpunct[, ]{(}{)}{,}{a}{}{,}%
\newcommand{\bProof}{\begin{proof}{Proof.}}
\newcommand{\eProof}{\hfill\Halmos\\  \end{proof}}

\usepackage{algorithm}
\usepackage{algorithmic}

\newcommand{\bea}{\begin{equation*}}
\newcommand{\eea}{\end{equation*}}

\usepackage[usenames,dvipsnames]{xcolor}

\newcommand{\gao}[1]{\textcolor{black}{{#1}}} 
\newcommand{\add}[1]{\textcolor{black}{{#1}}} 
\newcommand{\new}[1]{\textcolor{black}{{#1}}} 

\usepackage{caption}

\usepackage{comment}

\begin{document}

\begin{center}
		\Large \bf Square-root regret bounds for continuous-time episodic
Markov decision processes

	\end{center}
	\author{}
	\begin{center}
	{Xuefeng
			Gao}\,\footnote{Department of Systems
			Engineering and Engineering Management, The Chinese University of Hong Kong, Hong Kong, China.
			Email: \url{xfgao@se.cuhk.edu.hk.} },
		Xun Yu Zhou\,\footnote{Department of Industrial Engineering and Operations Research and The Data Science Institute, Columbia University, New York, NY 10027, USA. Email: \url{xz2574@columbia.edu.}
		}
	\end{center}
	
	\begin{center}
		\today
	\end{center}

\begin{abstract}
We study reinforcement learning for continuous-time Markov decision processes (MDPs) in the finite-horizon episodic setting. In contrast to discrete-time MDPs, the inter-transition times of a continuous-time MDP are exponentially distributed with rate parameters depending on the state--action pair at each transition.
We present a learning algorithm based on the methods of value iteration  and upper confidence bound. We derive an upper bound on the worst-case expected regret for the proposed algorithm, and establish a worst-case lower bound, with both bounds of the order of square-root on the number of episodes. Finally, we conduct simulation experiments to illustrate the performance of our algorithm.
\end{abstract}

\section{Introduction}

Reinforcement learning (RL) studies the problem of sequential decision making in an unknown environment by carefully balancing between exploration (learning) and exploitation (optimizing) \citep{sutton2018reinforcement}. While the RL study has a relatively long history, it has received considerable attention in the past decades due to the explosion of available data and rapid improvement of computing power.  A hitherto default mathematical framework  for RL is  Markov decision process (MDP), where the agent does not know the transition probabilities and can observe a reward resulting from an action but does not know the reward function itself.  There has been  extensive research on RL for \textit{discrete-time} MDPs (\add{DTMDPs}); see, e.g., \cite{Jaksch2010, osband2017posterior, azar2017minimax, jin2018q}. {However, much less attention has been paid to RL for \textit{continuous-time} MDPs, whereas  there are many real-world applications where one {\it needs} to interact with the unknown environment and learn the optimal strategies continuously in time.  Examples include autonomous driving, control of queueing
systems, control of infectious diseases, preventive maintenance and robot navigation;  see, e.g., \cite{guo2009continuous2, piunovskiy2020continuous}, Chapter 11 of \cite{puterman2014markov} and the references therein.}

In this paper we study RL for tabular continuous-time Markov decision processes (CTMDPs) in the finite-horizon, episodic setting, where an agent interacts with the unknown environment in episodes of a fixed length with finite state and action spaces.
The study of model-based (i.e. the underlying models are assumed to be known) finite-horizon CTMDPs  has a very long history, probably dating back to  \cite{miller1968finite}, with vast applications including queueing optimization \citep{lippman1976countable}, dynamic pricing \citep{gallego1994optimal}, and finance and insurance \citep{bauerle2011markov}. However, the problem of learning the optimal policy of a finite-horizon \textit{unknown} CTMDP has not been studied in the literature,  to our best knowledge.
The goal of this paper is to fill this gap by developing a learning algorithm for episodic RL in tabular CTMDPs with rigorous worst-case regret guarantees, as well as  providing a worst-case regret lower bound. Here, the regret measures the difference between the reward collected by the algorithm during learning and the reward of an optimal policy should the model parameters be completely known.

{At the outset, one may think learning CTMDPs is just a straightforward extension of its discrete-time counterpart in terms of theoretical analysis and algorithm design and thus not worthy of a separate study. This is not the case. The reason is because a CTMDP is characteristically different from a DTMDP in that, after an agent observes a state $x$ and takes an action $a$, the process remains in this state for a {\it random} holding time period that follows an exponential distribution with rate parameter $\lambda(x,a)$. Then the process jumps to another state $y$ with a transition probability $p(y|x,a)$.
For DTMDPs, the holding times (i.e., inter-transition times) are deterministic and take value of one time step in the general framework of semi-Markov decision processes \cite[Chapter 11]{puterman2014markov}.
This difference, along with the continuity of the time parameter,  leads to several subtle yet substantial challenges in the design and analysis of episodic RL algorithms for CTMDPs. 
In particular, existing derivations  of state-of-the-art worst-case regret bounds for learning tabular DTMDPs (see, e.g., \citealp{azar2017minimax, zanette2019tighter}) often rely on induction/recursion in the time parameter, which naturally  fails for  continuous-time problems.}

{In this paper, we develop a learning algorithm for CTMDPs with unknown transition probabilities and holding time rates, based on two key methodological ingredients.
First, we build on the value iteration algorithms of \cite{mamer1986successive, huang2011finite} for solving finite-horizon CTMDPs when all the model parameters are known.
This type of algorithms are based on an operator with a fixed point corresponding to the solution to the dynamic programming equation. A major benefit of this approach is that it naturally introduces a (discrete) number of iterations that can be used for induction arguments in regret analysis. 
Second, we take the idea of ``optimism under uncertainty", i.e., one acts as if the environment is as nice as
plausibly possible \citep{lattimore2020bandit}. This is a popular paradigm to address the exploration--exploitation trade-off in RL. To ensure optimism for the purpose of getting a reasonably good upper bound of the true value function, we carefully design the exploration bonus needed for efficient learning and regret analysis. }

{Because we are dealing with Markov chains (instead of, say, diffusion processes with continuous state space), it is only natural and indeed unavoidable that certain components of our analysis are inspired by those for discrete-time Markov chains, in particular the UCRL2 (Upper Confidence Reinforcement Learning) algorithm of \cite{Jaksch2010} and the UCBVI (Upper Confidence Bound Value Iteration) algorithm of \cite{azar2017minimax}, both originally developed  for RL in tabular DTMDPs.} At the beginning of each episode,
our algorithm estimates the transition probabilities  and the holding time rate using the past data/samples. Based on the level of uncertainty in these estimates, we carefully design an exploration bonus for each state--action pair. We then update the policy by applying a (modified) value iteration to the estimated CTMDP with the reward augmented by the bonus.

Our main result,  Theorem~\ref{thm:CT-UCB}, indicates  that the proposed RL algorithm has a worst case expected regret of $\tilde O (\sqrt{K})$ where $K$ is the number of episodes and $\tilde O$ ignores logarithmic factors and hides dependency on other constants. Moreover, Theorem~\ref{thm:lowerbound}  provides a lower bound
 showing that the square-root dependance of the worst case regret upper bound on $K$ is optimal (up to a logarithmic factor). Furthermore, we conduct simulation experiments to illustrate the performance of our algorithm.

{While our algorithm bears certain similarity to UCBVI of \cite{azar2017minimax} conceptually, there are {\it significant} differences in our algorithm design and regret analysis compared with the discrete time setting, which we elaborate below. }

\begin{itemize}

\item [(a)] {First, in contrast with learning DTMDPs, we need to estimate the holding time rate $\lambda(x,a)$ of a CTMDP for each state--action pair $(x,a)$ in our RL algorithm.   In the regret analysis, we also need to analyze the estimation error of $\lambda(x,a)$ by establishing a finite sample confidence bound.
Note that in the episodic setting, the random exponential holding time associated with each state--action pair may be truncated by the endpoint of the time horizon. This generates possibly non i.i.d. observations of holding times within the same episode. 
Hence, we can not directly apply concentration inequalities for i.i.d. exponential distributions to obtain the confidence bound. The key idea to overcome this difficulty is to `piece together' the samples of exponential holding times at a pair $(x,a)$ as realizations of inter-arrival times of a Poisson process with rate $\lambda(x,a)$ modelling the number of visits to the pair over time in an episode. Even if the time is eventually truncated, the number of visits to the pair $(x,a)$ (more precisely, those visits where the next states are observed) in an episode still follows a Poisson distribution conditioned on the total amount of time staying at $(x, a)$. Since the holding times at $(x,a)$ across different episodes are independent, we can then apply the tail bounds for Poisson distribution to derive the desired confidence bound for the holding time rate $\lambda(x,a)$. Such an analysis is not required for learning DTMDPs.
}

\item  [(b)]
{Second, because the value function of a CTMDP depends on both the transition probabilities and the holding time rates, the design of
the exploration bonus in our algorithm, which is used to bound estimation errors on the value function, is more sophisticated compared with the discrete time setting.  Specificaly,  we need to carefully analyze how the estimation errors of holding time rates and transition probabilities \textit{jointly} affect the computed value function in our learning algorithm. Moreover, because the ``Bellman operator'' in our continuous-time setting (see the operator $\mathcal{T}^a$ in Equation~\eqref{eq:Qa}) is more complicated due to the presence of exponential holding times, the error analysis of the learned model also become more involved. }

\item  [(c)]
{Third, our algorithm also differs from the classical UCBVI algorithm in terms of the planning oracle.
In constast to solving finite-horizon DTMDPs, the value iteration algorithm for solving \textit{finite-horizon} CTMDPs with known parameters converges generally in \textit{infinite} number of iterations. This leads to important difference in the regret analysis, in particular in the proof of optimism (see Lemma~\ref{lem:optimism}). In the study of learning DTMDPs, the optimism can be proved by backward induction on the time parameter. In our setting of episodic CTMDPs, however, induction is not enough to establish optimism. Extra analysis of (modified) value iterations for CTMDPs (Algorithm~\ref{alg:VI}) is required, because Algorithm~\ref{alg:VI} is stopped after a finite number of iterations and we need to quantify the error.  In this paper, we show that for \textit{finite-horizon} CTMDPs with bounded transition rates, the (modified) value iteration converges asymptotically with a linear rate $\rho \in (0,1)$ when the number of iterations approach infinity. This is achieved by showing that the Bellman operator for the empirical CTMDP with reward function augmented by the bonus (see the operator in Equation~\eqref{eq:Tak}) is a contraction mapping with the coefficient $\rho$. Such a technical analysis is not required for the classical UCBVI algorithm in the discrete-time setting. }

\item  [(d)]
{Fourth, in contrast to the  analysis of UCBVI for DTMDPs, the standard pigeon-hole argument (see, e.g., \cite{azar2017minimax} or Lemma 7.5 in Chapter 7 of \citealp{agarwal2019reinforcement}), which is used to bound the sum of bonuses evaluated on data to obtain sublinear regret, can \textit{not} be applied to our continuous time setting. Our exploration bonus involves both the estimation errors of transition probabilities and holding time rates. 
The estimation error of transition probabilities of a CTMDP is  \textit{not} inversely proportional to the square root of the \textit{total visit counts} at each state-action pair. This is because after visiting a state--action pair, the random holding time may be truncated by the endpoint of the horizon, and hence the next transition/state may not be observed. This subtle difference, compared with the discrete-time setting, leads to substantial challenges in our regret analysis. In particular, it is possible that the sum of estimation errors of transition probabilities evaluated on data is linear in $K$ for some sample paths (see \eqref{eq:sumK}). This shows that the standard pathwise pigeon-hole argument can not be directly used for obtaining {\it sublinear} regret in learning CTMDPs.
In our paper, we develop a new probabilistic method to overcome such a difficulty; see Proposition~\ref{prop:proberror} and its proof. 
On the other hand, the estimation error in the holding time rate of a CTMDP is inversely proportional to the square root of the total time spent at each state after an action, and this total spending time is \textit{real-valued} instead of \textit{integer-valued}. Hence, the pigeon-hole principle as a type of counting argument does not apply. {We overcome this difficulty by relating the total spending time with the visit counts for each state--action pair (see Proposition~\ref{prop:rateerror} and its proof)}.
}

\item  [(e)]
{ Finally,
the transitions of a finite-horizon CTMDP do not happen according to a Poisson process in general, because the rates of exponential holding times $\lambda(\cdot, \cdot)$ depend on the state--action pairs. As a consequence, while the horizon length of each episode is fixed, the number of transitions in different episodes are random variables that are generally unbounded and statistically different from each other. This is in sharp contrast with regret minimization in episodic DTMDPs, where the number of decision steps is often fixed and common across different episodes. In our regret analysis, we overcome this difficulty by `uniformizing' the total number of decision steps in each episode for CTMDPs with bounded transition rates, and then applying bounds for the maximum of $K$ independent Poisson distributions with the same rate. }

\end{itemize}

This paper is related to several studies on RL for CTMDPs in the \textit{infinite-horizon} setting.
Computational RL methods were developed for infinite-horizon CTMDPs as well as for  the more general semi-Markov decision processes (SMDPs) very early on  \citep{bradtke1995reinforcement, das1999solving}. However, available theoretical results on regret bounds for learning infinite-horizon CTMDPs are few and far between.
\cite{fruit2017exploration} study learning in continuous-time infinite-horizon average-reward SMDPs which is more general than CTMDPs in that the holding times can follow general distributions. They adapt the UCRL2 algorithm by \cite{Jaksch2010} and show that their algorithm achieves $\tilde O( \sqrt{ n})$ worst-case regret after $n$ decision steps. 
 Instead of worst-case (instance-independent) performance bounds, \cite{gao2022logarithmic} recently establish instance-dependent regret bounds that are logarithmic in time for learning tabular CTMDPs, again in the  infinite-horizon average-reward setting. It is well recognized that the finite-horizon continuous-time decision problem is different from and indeed generally  harder than its infinite-horizon counterpart in a number of aspects, including the need of considering the time variable in the value function and the truncation of the holding time discussed earlier. As such, our paper naturally differs from those on infinite-horizon problems.

We conclude this introduction by mentioning a growing body of literature on RL in continuous time with possibly continuous state and/or action spaces. Recently, \cite{wang2020reinforcement, jia2022policy, JZ22, JZ23} consider RL for diffusion processes and study respectively the problems of generating trial-and-error policies strategically, police evaluation, policy gradient and q-learning. However, the regret analysis remains largely untouched in that series of study. \cite{basei2021logarithmic, guo2021reinforcement} and \cite{szpruch2021exploration} study continuous-time RL for linear quadratic/convex models with {\it continuous} state spaces and propose algorithms with sublinear regret bounds in the finite-horizon episodic setting. By contrast, our work considers episodic RL for CTMDPs with a {\it discrete} state space.

The remainder of the paper is organized as follows. In Section~\ref{sec:episodicRL}, we formulate the problem of episodic RL for CTMDPs. In Section~\ref{sec:alg} we introduce our learning algorithm, while in Section~\ref{sec:result} we present the main results on regret bounds. Section~\ref{sec:simulation} reports the result of simulation experiments and
Section~\ref{sec:proof-mainresult} contains the proofs of the main results.
Finally Section~\ref{sec:conclusion} concludes. 

\section{Episodic RL in a Tabular CTMDP} \label{sec:episodicRL}

We consider a continuous-time Markov decision process (CTMDP) with a finite state space $\mathcal{S}$, a finite action space $\mathcal{A}$ and a finite time horizon $[0,H]$. The CTMDP evolves as follows \cite[Chapter 11]{puterman2014markov}. At the initial time $0,$ the process is in state $x_0 \in \mathcal{S} $ and an agent chooses an action $a_0 \in \mathcal{A}$. 
The process remains in state $x_0$ for a random holding time period $\tau_0$ that follows an exponential distribution with rate parameter $\lambda(x_0, a_0),$ while rewards are continuously accrued at a rate $r(x_0,a_0)$ during the holding period  $\tau_0$.
Then the process jumps to another state $x_1 \in \mathcal{S}$ at time $\tau_0$ with a transition probability $p(x_1 | x_0, a_0)$, and another action $a_1$ is made upon landing on $x_1$.  
This series of events is repeated until the end of the horizon, $H$. {It is immediate to see that if action $a$ is chosen in state $x$, then the joint probability that the holding time in state $x$ is not greater than $t $ and the next state is $y$ is given by $ \left(1- e^{-\lambda(x, a) t}\right) \cdot p (y | x,a)$. }

{For $(x,a) \in \mathcal{S} \times \mathcal{A}$, define the (controlled) transition rates
 \begin{align*}
q(y | x, a) := \lambda(x,a) \cdot p(y|x,a) \ge 0,  \quad \text{for $y \in \mathcal{S}, \; y \ne x$},
\end{align*}
and set $q(x | x, a) = - \lambda(x,a) \le 0$ so that $\sum_{z \in \mathcal{S}}q(z|x,a) \equiv 0.$
Mathematically, a finite-horizon CTMDP model $\mathcal{M}$ is characterized  by the following set: 
\begin{align}\label{eq:M}
\mathcal{M}: = (\mathcal{S}, \mathcal{A}, r(\cdot,\cdot), q(\cdot| \cdot,\cdot), H, x_0),
\end{align}
where $r(\cdot, \cdot)$ is the reward function, $q(\cdot| \cdot,\cdot)$ is the transition rate, $H<\infty$ is the length of the horizon and $x_0$ is the initial state of the CTMDP model.  Note that $q(\cdot| \cdot,\cdot)$ and $(\lambda(\cdot,\cdot), p(\cdot| \cdot,\cdot))$ are equivalent in the sense that they determine each other. Hence, we will also occasionally use 
\begin{align*}
\mathcal{M} = (\mathcal{S}, \mathcal{A}, r(\cdot,\cdot), \lambda(\cdot,\cdot), p(\cdot| \cdot,\cdot), H, x_0).
\end{align*} 
We emphasize that $(\lambda(x,a); (x,a) \in \mathcal{S} \times \mathcal{A})$ are the rate parameters of the exponential distributions of holding times in a CTMDP and they are {\it not} the discounting factors typically appearing in discrete-time reinforcement learning. }

Throughout this paper, we make the following assumption.
\begin{assumption}\label{eq:assume}
{There exist two constants $\lambda_{\min}, \lambda_{\max} \in (0,\infty)$, such that $\lambda(x,a) \in [\lambda_{\min}, \lambda_{\max}]$ for all $(x,a) \in \mathcal{S} \times \mathcal{A}$ and,  without loss of generality,  $r(x,a) \in [0,1]$ for all $(x,a) \in \mathcal{S} \times \mathcal{A}$. We also assume that the constant $\lambda_{\max}$ is known. }
\end{assumption}

{This assumption states that the holding time rates and reward rates are bounded, which is fairly natural since the state and action spaces are both finite. In addition, it is assumed that we know the specific value of  $\lambda_{\max}$, which is needed when we implement our proposed learning algorithm and carry out the theoretical analysis of the algorithm.  For some CTMDPs, one may be able to  obtain priori upper bounds on the holding time rates based on known information of the specific applied problems, although in general this may not be always the case. Finally, while our regret upper bound will depend on $\lambda_{\min}$, we do not need the knowledge of $\lambda_{\min}$ as an input for our algorithm. 
}

{
 Note that in our CTMDP model, the system evolves continuously in time, and the decision maker chooses actions at every transition when the system state changes/jumps.
Given a control policy $\pi,$ which is a map from $\mathcal{S} \times [0, H]$ to $\mathcal{A}$,  denote by $\tau_i$ the holding time after $i$-th jump of a CTMDP model, by
 $t_n: = \sum_{i=0}^{n-1} \tau_i$ the $n$-th jump time  with $t_0:=0$, by $x_n$ the state at $t_n,$ and by $a_n:=\pi(x_n, (H-t_n)^+)$ the action dictated by  $\pi$ at $t_n$, where $(H-t_n)^+$ denotes the remaining horizon.} 
We write
\begin{align*}
X(u): = \sum_{n \ge 0} I_{t_n \le u < t_{n+1}} x_n, \quad A(u): =  \sum_{n \ge 0} I_{t_n \le u < t_{n+1}} a_n, \quad u \ge 0,
\end{align*}
as the corresponding  continuous-time state and action processes respectively under $\pi,$ where $I_C$ denotes the indicator function on a set $C.$ Define the value function under $\pi$:
\begin{eqnarray} \label{eq:FH-control}
V^{\pi}(x,t): = \mathbb{E}^{\pi} \left[ \int_0^t r( X(u), A(u)) du \Big| X(0) =x \right], \quad (x, t) \in \mathcal{S} \times [0, H],
\end{eqnarray}
as well as the {\it optimal} value function (i.e. the maximum mean accumulated reward)
\begin{eqnarray} \label{eq:FH-control2}
V^*(x, t): = \sup_{\pi \in \Pi} V^{\pi}(x, t), \quad (x, t) \in \mathcal{S} \times [0, H],
\end{eqnarray}
where $\Pi$ denotes the set of all deterministic Markov policies. Because both the state and action spaces are finite, the existence of optimal policies is guaranteed under Assumption~\ref{eq:assume} \citep{mamer1986successive}.

Consider an agent who repeatedly interacts with an unknown finite-horizon CTMDP $\mathcal{M}$ over a total of $K$ episodes, where the time length of each episode is $H$. For simplicity we assume {the functional form of} the reward function $(r(x,a)))_{s \in \mathcal{S}, a \in \mathcal{A}}$ is known, but the transition probabilities $(p(\cdot |x, a))_{x \in \mathcal{S}, a \in \mathcal{A}}$ and the rates of exponential holding times $(\lambda(x,a))_{x \in \mathcal{S}, a \in \mathcal{A}}$  are unknown. Extending our algorithm and analysis in this paper to the case of unknown random rewards is not difficult; see, e.g., \cite{zanette2019tighter} for a standard  treatment.  
 In each episode $k=1, 2, \ldots, K,$ an arbitrary fixed initial state $x_0^k = x_0$ is picked for the CTMDP\footnote{The results of the paper can also be extended to the case where the initial states are drawn from a fixed distribution over $\mathcal{S}$.}. An algorithm $\textbf{algo}$ initializes and implements a policy $\pi_1$ for the first episode, and computes and executes  policy $\pi_k$ throughout episode $k$ based on past data up to the end of episode $k-1$, $k=2, \ldots, K$.  
The cumulative expected regret of $\textbf{algo}$ over $K$ episodes of interactions with $\mathcal{M}$ is defined by
\begin{align} \label{eq:regret}
\text{Regret}(\mathcal{M}, \textbf{algo}, K):= \mathbb{E} \left[ \sum_{k=1}^K V^{*}(x_0^k, H) -  \sum_{k= 1}^K  V^{\pi_k}(x_0^k, H)   \right].
\end{align}
The worst-case regret of $\textbf{algo}$ is defined by $\sup_{\mathcal{M \in \mathcal{C}}} \text{Regret}(\mathcal{M}, \textbf{algo}, K)$, where $\mathcal{C}$ is the set of all CTMDPs with $S$ states, $A$ actions, horizon $[0,H]$, and the holding-time rates and rewards satisfying  Assumption~\ref{eq:assume}.
Our objective is to obtain a bound on the worst-case regret that scales sublinearly in $K$ while having  polynomial dependance on the other parameters including $S, A, H$ and $\lambda_{\max}.$ \vspace{1mm}

\begin{remark}
{In this paper, we assume the transition probability $p(\cdot|\cdot,\cdot)$, reward function $r(\cdot,\cdot)$ and the holding time rate $\lambda(\cdot,\cdot)$ to be all time-invariant, i.e.,   the model is time homogeneous. This assumption simplifies the parameter estimation, algorithm development and regret analysis. For episodic learning of discrete-time tabular MDPs, extending the results from the homogeneous setting to nonhomogeneous one is straightforward.  For instance, one can still use the empirical transition probability to estimate the time-dependent transition probability $p_h(y|x,a)$ where $h=1, 2, \ldots, H$ for a discrete-time tabular MDP with horizon length $H$ steps. However, it is much more challenging to extend our current results to nonhomogeneous CTMDPs, precisely because the continuity of the time parameter. For illustrations, consider the time-dependent transition probability $p_t(y|x,a)$ with $t \in [0, H]$ for fixed $x,a, y$ in a CTMDP. This is an infinite-dimensional quantity due to $t \in [0, H]$. In addition, there are only finite (random) number of jumps in each episode of the CTMDP under Assumption~\ref{eq:assume}. This implies for almost every $t \in [0, H]$, one will not observe any transition occurred {\it exactly} at time $t$ in the episodic learning of CTMDPs with $K$ episodes.
Therefore, it is impossible to estimate $p_t(y|x,a)$ reliably for all $t \in [0, H]$, unless one imposes strong assumptions such as special structures of the function $p_t(y|x,a)$. This further illustrates the fundamental difference between episodic learning of CTMDPs and DTMDPs, even for model-based learning. Hence, in this paper we restrict our CTMDP model to the homogeneous setting.
Nevertheless, we believe considering episodic RL for nonhomogeneous CTMDP models is a very interesting topic worthy of future research.
}
\end{remark}

\subsection{Model-based dynamic programming approach}
Let us recall the dynamic programming approach for the model-based planning problem, i.e., solving the finite-horizon CTMDP \eqref{eq:FH-control2} when all the model parameters are known.

It has been established  that the optimal value function \eqref{eq:FH-control2} satisfies an optimality condition via the following Hamilton--Jacobi--Bellman (HJB) equation 
\begin{align} \label{eq:hjb-pde}
\frac{\partial v(x, t)  }{\partial t }  
&= \sup_{a \in \mathcal{A}} \left[ r(x,a) - \lambda(x, a) v( x, t ) + \lambda(x, a)  \sum_{z \in \mathcal{S}} v(z, t ) p(z| x, a )  \right], \quad (x,t) \in \mathcal{S} \times (0, H],\\
v(x, 0) &=0;  \nonumber
\end{align}
see, e.g., Theorem 1 in \cite{miller1968finite} or Proposition 4.1 in \cite{guo2015finite}. Note that this is a system of ordinary differential equations (ODEs) as the state space $\mathcal{S}$ is finitely discrete.
The model-based approch is to first solve \eqref{eq:hjb-pde} offline and then obtain the optimal policy by maximizing the right hand side of \eqref{eq:hjb-pde}. However, it is impossible to adapt this ODE approach to the learning setting because  \eqref{eq:hjb-pde} is unspecified when its coefficients are unknown. 
Instead, we choose an alternative optimality condition that gives rise  to
a value iteration algorithm for computing the optimal value function $V^*$.  This alternative  condition is equivalent to the HJB equation in the model-based setting; yet its idea can be carried over to the RL setting.


   To this end, we first introduce some notations. Denote by $\mathbb{M}$  the Banach space of all real-valued measurable functions $u: \mathcal{S} \times [0, H] \rightarrow [0, \infty)$  satisfying
   $||u||_{\infty}:= \sup_{(x, t) \in \mathcal{S} \times [0, H]} u (x,t) < \infty$ and $u(x, 0) \equiv 0$ for all $x \in \mathcal{S}$.
    Given a CTMDP $\mathcal{M}$ in \eqref{eq:M}, we define two associated operators $\mathcal{T}^a$ and  $G$ from $\mathbb{M}$ to $\mathbb{M}$ as follows: for $u\in \mathbb{M}$, $(x, t) \in \mathcal{S} \times [0, H] $ and $a \in \mathcal{A}$,
    \begin{align} \label{eq:Qa}
    (\mathcal{T}^a u) (x,t)  & := r(x,a) \cdot \int_0^te^{-\lambda(x,a) s} ds+ \sum_{y} \int_0^t\lambda(x,a) e^{-\lambda(x,a) s}  u (y, t-s) p(y|x,a) ds  \nonumber \\
    & = r(x,a )   \cdot  \mathbb{E}_{ \tau \sim \exp(\lambda(x,a))} [t \wedge \tau] + \mathbb{E}_{ \tau \sim \exp(\lambda(x,a)), Y \sim p(\cdot|x,a)} [ u (Y, (t-\tau)^+)] ,
    \end{align}
    where $t \wedge \tau = \min\{t, \tau\}$ and $\exp(\lambda)$ denotes an exponential distribution with rate $\lambda,$
     and
    \begin{align*} 
    (G u) (x,t) &:= \sup_{a \in \mathcal{A}} (\mathcal{T}^a u) (x,t).
    \end{align*}

    Then we have the following results; see Theorem 1 in \cite{mamer1986successive} and Theorems~3.1 and 3.2 in \cite{huang2011finite}.

\begin{itemize}
\item (Optimality equation) The optimal value function $V^*$ in \eqref{eq:FH-control2} is the unique solution in $\mathbb{M}$ to the Bellman optimality equation $V^*= G V^*.$
\item (Verification) The function $ a^*(x, t):  = \arg \max_{a \in \mathcal{A}} (\mathcal{T}^a V^*) (x,t)$, that maps a state--time pair to an action, is an optimal policy for \eqref{eq:FH-control2}.
\item (Value iteration)  Let $V_0^*:=0$ and $V_{n+1}^*: = G V_n^*$. Then $V_{n+1}^* \ge V_n^*$ and  $\lim_{n \rightarrow \infty} ||V^*_n - V^*||_{\infty}=0.$
\end{itemize}

Intuitively, one can view
$V_n^*$ as the optimal value function of a modified control problem in
which the objective is to maximize the expected total reward from
the first $n$ transitions or up to time $H$, whichever
comes first \citep{mamer1986successive}.
The results above will play an essential role in the design and analysis of our RL algorithm for episodic CTMDPs. In particular, the value iteration naturally introduces a sequence of numbers -- those of the iterations -- and makes it possible to apply induction arguments to the continuous-time setting; such induction arguments have been critical in regret analysis on  episodic RL algorithms for discrete-time tabular MDPs.

\subsection{Issues with upfront time discretization}

{Before we proceed, let us discuss an ``obvious"  approach for treating a CTMDP -- one that  first discretizes time and then applies existing RL algorithms to the resulting DTMDP. There are several issues with this approach as we discuss below. }

{ Consider a uniform time discretization of the horizon $[0, H]$ with discretization step size $\Delta>0$. For the simplicity of discussion, we assume $H/\Delta$ is an integer. Then the CTMDP model is approximated by a DTMDP:
\begin{eqnarray} \label{eq:FH-control-discrete}
\sup_{\pi }V_{\Delta}^{\pi}(x, H) : = \mathbb{E}^{\pi} \left[ \sum_{j=0}^{H/\Delta-1} r( X({j\Delta}), A({j\Delta}))   | X(0) = x\right] \cdot \Delta,
\end{eqnarray}
where $r(x,a) \in [0,1]$ for all $x,a,$ and the value function $V_{\Delta}^{\pi}(x, H)$ approximates \eqref{eq:FH-control} for the CTMDP when $\Delta$ is small. Importantly, the number of decision steps in this DTMDP is $\frac{H}{\Delta},$ while in the original CTMDP the expected number of decision steps is only $O(H)$ due to Assumption~\ref{eq:assume}.
One can apply existing learning algorithms for DTMDP to this model, which however will lead to vacuous regret bounds as $\Delta \rightarrow 0.$ To see this, consider for example the UCBVI--BF algorithm with Bernstein bonus \citep{azar2017minimax}, which has a regret bound $\tilde O(\bar H \sqrt{SAK} + \bar H^2 S^2 A + \bar H^{1.5} \sqrt{K} )$ for a DTMDP with $\bar H$ decision steps and stationary transitions, where $K$ is the number of episodes. This bound matches the regret lower bound for learning tabular DTMDP when $K \ge \bar H^2 S^3 A$ and $SA>H$ \citep{domingues2021episodic}.
Applying UCBVI--BF to the episodic learning of DTMDP \eqref{eq:FH-control-discrete}, we have
\begin{align*}
\text{Regret}^{\Delta}&:= \mathbb{E} \left[ K V_{\Delta}^{*}(x_0, H) -  \sum_{k= 0}^{K-1} V_{\Delta}^{\pi_k}(x_0, H)   \right] \\
&=  \Delta \cdot \tilde O \left( \frac{H}{\Delta} \sqrt{SAK} + \left(  \frac{H}{\Delta} \right)^2 S^2 A + \left(  \frac{H}{\Delta} \right)^{1.5} \sqrt{K} \right),
\end{align*}
where $V_{\Delta}^{*}$ is the optimal value function for the DTMDP \eqref{eq:FH-control-discrete} and $\pi_k$ is the executed policy in episode $k$ under the UCBVI--BF algorithm.
It is clear that the bound for $\text{Regret}^{\Delta}$ grows to $\infty$ when $\Delta \rightarrow 0.$ Considering alternative learning algorithms such as EULER in \cite{zanette2019tighter} for DTMDPs leads to a similar explosion of regret bound.
}

{Another issue with the upfront discretization approach is that the error bound between the value function of the CTMDP and that of the  DTMDP from time-discretization is in general unknown, i.e., it is hard to compute $V^{\pi} - V_{\Delta}^{\pi}$ with explicit constants. Hence, while one may expect that $\text{Regret}^{\Delta}$ converges to the regret of the CTMDP model when $\Delta \rightarrow 0$, one is generally unable to quantify the gap explicitly so as to give information in $S, A, H$ and the bounds on the holding time rates $\lambda_{\max}$ and $\lambda_{\min}$. }

{Finally, from the implementation perspective, it is challenging to select the discretization step size $\Delta$ properly. Indeed, it is well known in the RL community that the performance of RL algorithms can be very sensitive with respect to discreteization step size; see, e.g., \cite{tallec2019making} in which it is empirically shown that standard $Q-$learning methods are not robust to changes in time discretization of continuous-time control problems.}


\section{The CT-UCBVI Algorithm} \label{sec:alg}

In this section, we introduce an RL algorithm, named the CT-UCBVI (continuous-time upper confidence bound value iteration) algorithm,  for learning episodic tabular CTMDPs. It is a continuous-time analogue of the UCBVI algorithm developed in \cite{azar2017minimax} for learning discrete-time MDPs.

{The CT-UCBVI algorithm updates the estimates of model parameters $(\lambda(\cdot,\cdot), p(\cdot| \cdot,\cdot))$ at the beginning of each episode. 
 Let $(x_n^l, a_n^l, \tau_n^l)_{n}$ be the trajectories of the state, action and holding time in episode $l$ of the CTMDP, and $t_n^l = \sum_{i=0}^{n-1} \tau_i^l$ where the discrete index $n$ counts the number of jumps/transitions occurred in this episode; see Section 2.}
 A state--action--state triplet $(x,a,y)$ means that the process is in $x$, takes an action $a$ and then moves to $y$. Similarly, a state--action pair $(x,a)$ signifies that the process is in $x$ and  takes an action $a$.

{ {At the start of episode $k$,  we set the accumulated duration at  $(x,a)$ up to the end of episode $k-1$ as $T^{k}(x,a)$, for $x \in \mathcal{S}$ and $a \in \mathcal{A}$. Mathematically, it is given by 
\begin{align}
T^{k}(x,a) &= \sum_{n=0} ^{\infty}\sum_{l=1}^{k-1} \tau_n^l \cdot 1_{ \{( x_n^l, a_n ^l) = (x,a)\}} \cdot 1_{t_n^l \le H}. \label{eq:Timek}
\end{align} }
Here, the indicator function on the event $\{t_n^l \le H\}$ ensures that  the final jumping time of the CTMDP in episode $l$ does not exceed the end of the horizon, $H.$ Note that the observation of the `last' holding time in each episode may be truncated by $H$, i.e., if $t_n^l \le H < t_{n+1}^l$, then the actual observed holding time is $\tau_n^l = H - t_n^l$ instead of $ t_{n+1}^l - t_n^l$.
In this case, the truncated $\tau_n^l$ is {\it not}  a sample of an exponential distribution, noting we have slightly abused the notation for $\tau_n^l$ here.}

{We also set
the observed accumulated visit counts to $(x,a,y)$  up to the end of episode $k-1$ as $N^k(x,a, y)$ for $x, y \in \mathcal{S}$ and $a \in \mathcal{A}$. Note that $x_{n+1}^l$ is {\it not} observed if $t_n^l < H< t_{n+1}^l.$\footnote{This seemingly innocent subtlety actually causes a substantial difficulty in our regret analysis below.} {Hence, for the purpose of estimating parameters of the CTMDP, we also introduce $N^{k, +}(x,a)$, which differs from from the cumulative visit counts to $(x,a)$ up to the end of episode $k-1$ in that does not include the visit count to $(x,a)$ if the next state is not observed due to the finite horizon.} Specifically, we let
\begin{align}
N^k(x,a, y) & = \sum_{n=0} ^{\infty} \sum_{l=1}^{k-1} 1_{ \{( x_n^l, a_n^l, x_{n+1}^l) = (x,a , y)\}} \cdot 1_{t_{n+1}^l \le H}, \label{eq:xay}\\
N^{k, +}(x,a) &= \sum_{n=0} ^{\infty}\sum_{l=1}^{k-1} 1_{ \{( x_n^l, a_n ^l) = (x,a)\}} \cdot 1_{t_{n+1}^l \le H}.   \label{eq:Nkplus}
\end{align}  }

At the beginning of episode $k,$ we  compute the empirical transition probabilities by
\begin{align} \label{eq:empirical-P}
{ \hat p_k (y|x, a): = \frac{N^k(x,a, y)}{\max\{1,N^{k, +}(x,a) \}}, }
\end{align}
as well as an estimator for the holding-time rate $\lambda(x,a)$ by
\begin{align} \label{eq:rate-est}
\hat \lambda_k(x,a):= \min \left\{ \frac{{N^{k, +}(x,a)}}{T^k(x,a)}, \lambda_{\max} \right\}.
\end{align}
If $(x,a)$ has not been sampled prior to  episode $k$, we define $\hat p_k (y |x, a) = 0$ and $\hat \lambda_k(x,a) =0$  for all $y \in \mathcal{S}$.    \footnote{Strictly speaking, $\hat p_k (\cdot |x, a)$ is not a probability vector on $\mathcal{S}$ in this case. However, this will not affect our analysis  because we set $\hat \lambda_k(x,a) =0$ when $(x,a)$ has not been sampled before, see \eqref{eq:Tak}. } In particular, when $k=1,$ there is no data available so $\hat p_1 (y|x, a) = 0$ and $\hat \lambda_1(x,a) =0$.
 Note that we truncate the empirical holding-time rate $ \frac{\new{N^{k, +}(x,a)}}{T^k(x,a)}$ by $\lambda_{\max}$ -- while this is natural under our assumption  $\lambda(x,a) \le \lambda_{\max}$, the truncation turns out also important in
our regret analysis below.

The CT-UCBVI algorithm encourages exploration by awarding some `bonus' for exploring certain state--action pairs  in learning CTMDPs. The bonus, which is a function of $(x,a)$,  is constructed based on the level of uncertainty in learned transition probabilities \eqref{eq:empirical-P} and holding time rates \eqref{eq:rate-est} to incentivize the agent to explore those  pairs with high uncertainty.
Mathematically,
the  bonus function $b^k$ for each episode $k$  is defined by
\begin{align} \label{eq:bonus-k}
b^k(x,a,\delta) = \left(\frac{\lambda_{\max}}{1-e^{-\lambda_{\max} H}} \vee 1\right) \cdot   \left({H^2}  \sqrt{\frac{  \lambda_{\max} L(\delta) }{   \max\{T^k(x,a),  L(\delta)/ \lambda_{\max} \}}} + H \sqrt{\frac{2 [ S \ln 2 + \ln\left(SA H K^2 /\delta \right)]}{ \max\{1, \add{ N^{k, +}(x,a) }\}} }  \right),
\end{align}
where  $(x,a) \in \mathcal{S} \times \mathcal{A}$,  
$\delta \in (0,1)$ is an input parameter in our learning algorithm, and $L(\delta):= 4 \ln (2SAK/\delta)$. 

{Let us provide some explanations on the expression of the bonus $b^k(x,a,\delta)$. In \eqref{eq:bonus-k}, the term $\sqrt{\frac{  \lambda_{\max} L(\delta) }{   \max\{T^k(x,a),  L(\delta)/ \lambda_{\max} \}}} := \beta_k(x,a, \delta)$ is the confidence bound for the holding time rate (see Lemma \ref{lem:rateUCB} in Section~\ref{sec:pre-lemma}), and the term $\sqrt{\frac{2 [ S \ln 2 + \ln\left(SA H K^2 /\delta \right)]}{ \max\{1, N^{k, +}(x,a)\}}}:=  \alpha_k(x,a, \delta)$ is the confidence bound for the transition probabilities (see Lemma \ref{lem:CI-p} in Section~\ref{sec:pre-lemma}). By setting the bonus $b^k(x,a,\delta)$ as in \eqref{eq:bonus-k},
one can ensure optimism in regret analysis, i.e., the computed value function from our algorithm is a high probability upper bound of the true optimal value function (up to some error). See Lemma~\ref{lem:optimism} in Section~\ref{sec:pre-lemma} for further details. }

The learning algorithm CT-UCBVI, presented as Algorithm~\ref{alg: CTUCB},  proceeds as follows. In episode $k=1, \ldots, K,$ the algorithm  uses the past data (trajectories of state, action and holding time) up to the end of episode $k-1$ to estimate the parameters of the CTMDP, and to compute the reward bonus function $b^k$. Then it applies the modified value iteration procedure (Algorithm~\ref{alg:VI}) to the estimated model with a combined reward function  $r + b^k$, and executes the returned policy $\pi^k$ from the value iteration in episode $k$. Finally, it observes the data collected in episode $k$ and updates the data for model estimation  in the next episode. The algorithm repeats the above procedure for $K$ episodes.

Meanwhile, the subroutine, Algorithm~\ref{alg:VI}, is adapted from the value iteration scheme developed in \cite{huang2011finite}. The main modification needed here  is that we truncate the value function by $t$ in the algorithm.  Note that because the reward $r(x,a) \in [0,1]$, the value $V^{\pi}(x, t)$ of any policy $\pi$ under the true CTMDP model is bounded by $t$.
Such a truncation trick is standard in the discrete-time RL literature (see, e.g., \citealp{azar2017minimax, jin2018q}) which facilitates the regret analysis.
\vspace{2mm}

\newpage
    \begin{algorithm}[!ht]
        \caption{The CT-UCBVI Algorithm}
        \label{alg: CTUCB}
        \begin{algorithmic}[1]
            \REQUIRE Parameters $\delta, \epsilon_k \in (0,1)$ for $k \ge 1$, $\lambda_{\max}, \mathcal{S}, \mathcal{A}, H$ and reward function $r$.
            \STATE Initialization: set initial state $x_0;$
            \FOR {episode $k=1,2,\ldots, K$}
          \STATE Past data $\mathcal{D}= \{ (x_n^l, a_n^l, \tau_n^l)_{ n \ge 0}: l <k \} $;
            \STATE
             For all $(x, a, y) \in \mathcal{S} \times \mathcal{A} \times \mathcal{S}$: \\
            \STATE   Compute the empirical transition probabilities  $\hat p_k(y|x,a)$ by \eqref{eq:empirical-P}; \\
             \STATE   Compute the estimated holding-time rate $\hat \lambda_k(x,a)$ by \eqref{eq:rate-est}; \\
            \STATE   Compute the reward bonus $b^k(x,a, \delta)$ by \eqref{eq:bonus-k}; \\
%
%

\STATE  Apply modified value iteration (Algorithm~\ref{alg:VI}) to the CTMDP model $(\mathcal{S}, \mathcal{A}, r(\cdot,\cdot) + b^k(\cdot,\cdot,\delta), \hat \lambda_k(\cdot,\cdot), \hat p_k(\cdot|\cdot,\cdot), H, {x_0})$ with accuracy parameter $\epsilon_k$ and return value function $ \hat V^{k}_{n_k}$;

  \STATE Compute  policy $\pi^k$:
    \begin{align}\label{eq:pi-k}
\pi^k( x, t )   =  \arg \max_{a \in \mathcal{A}} \{ \mathcal{T}^{a,k} \hat V_{n_k}^k (x,t) \}, \quad \text{for all $ (x, t)\in \mathcal{S} \times [0, H]$,}
\end{align}
where
    \begin{align} \label{eq:Tak}
    \mathcal{T}^{a,k}  \hat V^{k}_{n_k} (x,t)  & =[ r(x,a) + b^k(x,a, \delta) ]\cdot \int_0^te^{-\hat \lambda_k (x,a) s} ds\nonumber  \\
    & \quad + \sum_{y} \int_0^t \hat \lambda_k (x,a) e^{-\hat \lambda_k(x,a) s}   \hat V^{k}_{n_k} (y, t-s) \hat p_k(y|x,a) ds;
    \end{align}
 \STATE Execute $\pi^k$ in episode $k$;

 \STATE Observe data $(x_n^k, a_n^k, \tau_n^k)_{ n \ge 0}$ up to time $H$; 
        \ENDFOR
        \end{algorithmic}
    \end{algorithm}

\vspace{1cm}

    \begin{algorithm}[!ht]
        \caption{Modified Value Iteration for CTMDP}
        \label{alg:VI}
        \begin{algorithmic}[1]
            \REQUIRE A CTMDP model $(\mathcal{S}, \mathcal{A}, r(\cdot,\cdot), \lambda(\cdot,\cdot), p(\cdot|\cdot,\cdot), H, x_0)$ and accuracy paremeter $\epsilon.$
            \STATE Initialization: Set $n=0$ and $ V_n (x,t)=0$ for all $ (x, t)\in \mathcal{S} \times [0, H]$;
           \STATE  Repeat \\

            \STATE \quad $n = n+1$;\\
            \STATE   \quad Compute 
\add{
\begin{align} \label{eq:m-VI}
 V_{n+1} (x,t) &= \min\{ t, \max_{a \in \mathcal{A}} \mathcal{T}^a V_n(x,t) \}, \quad \text{for all $ (x, t)\in \mathcal{S} \times [0, H]$};
\end{align} }
 \\
             \STATE   Until  $||V_{n} - V_{n-1}||_{\infty} < \epsilon$;\\
            \STATE   Return $V_n$.

        \end{algorithmic}
    \end{algorithm}
\newpage

 \begin{remark}
    In Step 8 of the CT-UCBVI algorithm,  we apply the modified value iteration to solve the CTMDP model $(\mathcal{S}, \mathcal{A}, r(\cdot,\cdot) + b^k(\cdot,\cdot,\delta), \hat \lambda_k(\cdot,\cdot), \hat p_k(\cdot|\cdot,\cdot), H, {x_0})$ approximately. An alternative method  might be to time-discretize the HJB equation~\eqref{eq:hjb-pde} and solve it numerically to obtain an approximated value function. The problem with this approach for regret analysis, however, is that {the error resulting from time--discretization is unknown. }
        \end{remark}

 \begin{remark}
{In the CT-UCBVI algorithm, we assume that the type of integral in \eqref{eq:Tak} can be explicitly computed for every $t \in [0, H]$. Then the policy $\pi^k(x,t)$ can be computed by \eqref{eq:pi-k} for all $(x,t) \in \mathcal{S} \times [0, H]$, because the state and action spaces are both finite. In our actual algorithm implementations in the simulation experiments, we discretize the integral and first compute $\pi^k(x,t_i)$ for $x \in \mathcal{S}$ where $0 =t_0 < t_1 < \ldots < t_N =H.$ Given a state $x,$ to compute $\pi^k(x, t)$ for all $t \in [0, H]$, we then use piecewise-constant interpolation. The approximation error can be made arbitrarily small with a small grid size.
}
\end{remark}

\section{Main Results}\label{sec:result}

  Recall that $\mathcal{C}$ denote the set of all CTMDPs with $S$ states, $A$ actions, horizon $[0,H]$, and the holding-time rates and rewards satisfying  Assumption~\ref{eq:assume}.  The main result of the paper is the following theorem, whose proof  is delayed to Section~\ref{sec:proof-mainresult}.
\begin{theorem}\label{thm:CT-UCB}
The CT-UCBVI algorithm achieves the following worst-case regret bound:
{
\begin{align} \label{eq:regret-bound}
&\sup_{\mathcal{M \in \mathcal{C}}} \text{Regret}(\mathcal{M}, \text{CT-UCBVI}, K) \nonumber \\
& \le 3 \left(   \left(\frac{\lambda_{\max}}{1-e^{-\lambda_{\max} H}} \vee 1\right) H+1\right) \cdot \sqrt{SA K} \cdot  \left( H+  \frac{1}{\lambda_{\min} }\right) \cdot   (\lambda_{\max} H +1)  \frac{\ln K}{ \ln(\ln(K) +1)}  \nonumber \\
& \quad   \cdot \left(   \sqrt{ 2 S + 6\ln (2SA K H) }  + 4 \lambda_{\max} H  \sqrt{ \ln(2SAKH)} \right) \nonumber  \\
& \quad \quad  + \sum_{k= 1}^{K} {\epsilon_k} \cdot e^{\lambda_{\max} H }+ 1.
\end{align} }
where $\epsilon_k$, $k=1,2,\cdots,K$, are the accuracy parameters given in Algorithm~\ref{alg: CTUCB}.
\end{theorem}

{Theorem~\ref{thm:CT-UCB} immediately implies the following result.
\begin{corollary}
If $\lambda_{\max}, H >1$ and $\epsilon_k = 1/k^{\alpha} \cdot  e^{- \lambda_{\max} H}$ for $\alpha \ge 1/2$, then
{\begin{align} \label{eq:regret-b1}
\sup_{\mathcal{M \in \mathcal{C}}} \text{Regret}(\mathcal{M}, \text{CT-UCBVI}, K) = \tilde O \left( \lambda_{\max}^{2} H^2  \left( H+  \frac{1}{\lambda_{\min} }\right)  \sqrt{SA K} \left[  \sqrt{S}   +  \lambda_{\max} H \right] \right) ,
\end{align} }
where the notation $\tilde O$ ignores the logarithmic factors in $H, S, A$ and $K$.
\end{corollary}
}

\vspace{2mm}

Theorem~\ref{thm:CT-UCB} is complemented by Theorem~\ref{thm:lowerbound} below, which gives a regret lower bound.
We first state the following assumption.
\begin{assumption}\label{assume:LB}
The number of states and actions satisfy $S \ge 6, A \ge 2$, and there exists an integer $d$ such that $S = 2 + \frac{A^d - 1}{A-1}$.
\end{assumption}

 Assumption~\ref{assume:LB} is adapted from \cite{domingues2021episodic} (see Assumption 1 therein), where a rigorous proof of the minimax regret lower bound for episodic RL in discrete-time tabular MDPs is provided. \footnote{Assumption 1 of \cite{domingues2021episodic} requires $S = 3 + \frac{A^d - 1}{A-1}$ because  non-homogeneous transitions are considered there, as well as $H \ge 3d$ which is not needed in our analysis. } Note that $S = 2 + \frac{A^d - 1}{A-1} \equiv  2 + \sum_{i=0}^{d-1}A^i$. Hence, one can use $S-2$ states to construct a \textit{full A-ary tree} of depth $d-1$. In this rooted tree, each node has exactly $A$ children and the total number of nodes is given by $\sum_{i=0}^{d-1}A^i = S-2$. This tree construction allows one to design difficult CTMDP instances that are needed for proving the regret lower bound.
 Assumption~\ref{assume:LB} is imposed here for the purpose of a cleaner  analysis, and it can be relaxed following the discussion in Appendix D of \cite{domingues2021episodic}.

The proof of the following result is given in Section~\ref{sec:proof-mainresult}.

\begin{theorem}\label{thm:lowerbound}
Under Assumption~\ref{assume:LB}, for any algorithm \textbf{algo} there exists an $\mathcal{M \in \mathcal{C}}$  such that
\begin{align} \label{eq:regret-lowerbound}
\text{Regret}(\mathcal{M}, \textbf{algo}, K) \ge \frac{1}{12 \sqrt{2}} \cdot \mathbb{E} [ (H - \gamma_{d})^+]  \sqrt{SAK},\;\;\mbox{ for all } K \ge SA/2,
\end{align}
where $\gamma_d$ is a random variable following an Erlang distribution with rate parameter $\lambda_{\max}$ and shape parameter $d$. Moreover, if $d< (1- c_2) \lambda_{\max} H$ for some universal constant $c_2 \in (0,1)$, then
\begin{align} \label{eq:regret-lowerbound2}
\text{Regret}(\mathcal{M}, \textbf{algo}, K) \ge  \frac{c_2}{12 \sqrt{2}} \cdot H  \sqrt{SAK}.
\end{align}
\end{theorem}

The regret upper bound in Theorem~\ref{thm:CT-UCB} and the lower bound in Theorem~\ref{thm:lowerbound} suggest that our proposed CT-UCBVI algorithm achieves a regret rate with the optimal dependence on the number of episodes $K$ as well as on the number of actions $A$, up to logarithmic factors. While the bounds on $K$ are the most important as they inform the convergence rates of learning algorithms, it remains  a significant  open question whether one can improve the dependence of these bounds on $S, H$ and $\lambda_{\max}$  to narrow down the gap between the upper and lower bounds. For episodic RL in discrete-time tabular MDPs with stationary transitions, the UCBVI--BF algorithm (with Bernstein's bonuses) in \cite{azar2017minimax} has a regret bound of $\tilde O(H  \sqrt{SAK} + H^2S^2 A + H \sqrt{HK}) $. When $K \ge H^2 S^3 A$ and $SA \ge H$, the leading term in their regret bound is $\tilde O(H  \sqrt{SAK}) $, which matches the established lower bound of $\Omega(H  \sqrt{SAK})$  for discrete-time episodic MDPs,  up to logarithmic factors \citep{domingues2021episodic}. Our upper bound in Theorem~\ref{thm:CT-UCB} scales with $S$ linearly. Compared with the leading term $\tilde O(H  \sqrt{SAK}) $ in \cite{azar2017minimax},
the extra $\sqrt{S}$ factor comes from the fact that we apply concentration inequalities to the transition probability vector which is  $S$-dimensional; see Lemma~\ref{lem:CI-p}. \cite{azar2017minimax} instead maintain confidence intervals on the optimal value function directly; see also \cite{dann2019policy, zanette2019tighter} for similar approaches. The primary difficulty in extending the method of \cite{azar2017minimax} to our continuous-time setting is that the number of decision steps in each episode is random and different, and unbounded in general.  In addition, due to the inherent random holding times, our regret upper bound has a worse dependance on $H$ compared with the discrete-time counterparts.
\cite{azar2017minimax} consider Bernstein-type ``exploration bonuses" that use the empirical variance of the estimated value function at the next state to achieve tight (linear) dependance on $H$ (in the leading term of their regret bound).
It is an open problem how to extend their idea and  technique to our setting in order to  improve the dependancy of the upper bound on $H$.

\vspace{2mm}
\begin{remark}
If we assume the constant $\lambda_{\min}>0$ in Assumption~\ref{eq:assume} is also known to the agent, then we may improve the $H$ dependance in the regret upper bound. Specifically, we can truncate the empirical holding time rate \eqref{eq:rate-est} so that the new estimator $\hat \lambda_k (x,a) \ge \lambda_{\min}$ for all $(x,a)$. Then  in the proof of Lemma~\ref{lem:v-v-continuoustime}, the integral $\int_0^{(H-t_m^k)^+} e^{-\hat \lambda_k(x_0^k, a_0^k) s} ds$ (c.f. \eqref{eq:reward-disc}) can be upper bounded by $\frac{1}{\lambda_{\min}} (1 - e^{ -\lambda_{\min} (H-t_m^k)^+})$, instead of the quantity $(H-t_m^k)^+.$ One can then verify that when $\lambda_{\max}, H >1$,
the regret bound in \eqref{eq:regret-b1} can be improved to
\begin{align*} 
 \tilde O \left(   \frac{\lambda_{\max}^{2} H^2}{\lambda_{\min} }  \sqrt{SA K} \left[  \sqrt{S}   +  \lambda_{\max} H \right] \right) .
\end{align*} 
\end{remark}


\section{Simulation Results} \label{sec:simulation}
In this section, we demonstrate the performance of the CT-UCBVI algorithm by simuations. As there are no existing benchmarks for episodic RL in CTMDPs to the best of our knowledge,  we will take a continuous-time machine operation and repair problem taken from \cite{puterman2014markov}.

We first discuss the numerical computation of the expected regret in \eqref{eq:regret} given a CTMDP. First, we need to compute $V^{*}(x_0, H)$, the optimal value function of the CTMDP when the model parameters are known. This can be done by applying the value iteration procedure in Algorithm \ref{alg:VI}, except that there is no need to consider the truncation by $t$ in \eqref{eq:m-VI} in this case because it holds automatically that $V^{*}(x_0, t) \le t$ for all $t \in [0, H]$. Second, we need to numerically compute $V^{\pi_k}(x_0, H)$, which is the cumulative reward collected up to time $H$ in the known environment under the policy $\pi^k$ for each episode $k=1,2 ,\ldots, K.$ This is a problem of policy evaluation for the CTMDP. Lemma 3.1 in \cite{huang2011finite} yileds that
\begin{align*}
V^{\pi_k}(x, t) = \mathcal{T}^{\pi^k} V^{\pi_k} (x,t), \quad \text{for all $ (x, t)\in \mathcal{S} \times [0, H]$,}
\end{align*}
where, with a slight abuse of notation (compare with \eqref{eq:Qa}), the operator $\mathcal{T}^{\pi^k}$ is defined by
    \begin{align*}
    \mathcal{T}^{\pi^k}  u (x,t)  & = r(x, \pi_k(x,t)) \cdot \int_0^te^{-\lambda(x, \pi_k(x,t)) s} ds \\
    & \quad + \sum_{y} \int_0^t\lambda(x,\pi_k(x,t)) e^{-\lambda(x,\pi_k(x,t)) s}  u (y, t-s) p(y|x,\pi_k(x,t)) ds.
    \end{align*}
Hence we can apply the following iterative scheme to compute $V^{\pi_k}(x, t)$ for $ (x, t)\in \mathcal{S} \times [0, H]$:
\begin{itemize}
\item Set $u_0(x, t)=0$ for all $ (x, t)\in \mathcal{S} \times [0, H]$;
\item Compute
\begin{align*}
u_{n+1} (x, t)=   \mathcal{T}^{\pi^k}  u_n(x,t), \quad \text{for $ (x, t)\in \mathcal{S} \times [0, H]$,}
\end{align*}
 until $||u_{n+1} - u_{n}||_{\infty} < \epsilon$ for some pre-specified accuracy parameter $\epsilon;$
\item Return $u_n$ as an approximation of $V^{\pi_k}.$
\end{itemize}
Once we know how to compute $V^{*}$ and $V^{\pi_k}$, we can then compute the expected regret in \eqref{eq:regret} by averaging the regret over multiple independent runs.


We take Problem 11.2 of \cite{puterman2014markov} for our simulation experiment. A machine is either in state 0 representing it is operating, or in state 1 signifying it is under repair. In each state, the decision maker can choose an action from $\mathcal{A}: = \{ \text{slow, fast}\}\equiv \{a_1,a_2\} $.  Specifically, if the machine is in state 0 and $a_1$ is chosen, then the machine remains in the operating mode for an exponentially distributed amount of time with the failure rate $\lambda(0, a_1) = 3$ and the reward rate  $r(0, a_1) =5$; if $a_2$ is chosen, then the failure rate is $\lambda(0, a_2) = 5$ and the reward rate $r(0, a_2) =8$. Hence, when the machine operates at the fast rate, it yields higher reward but breaks down more frequently than when it operates at the slow rate. On the other hand, when the machine is in state 1 and action $a_1$ is chosen, then the repair time distribution is exponential with the recovery  rate
$\lambda(1 ,a_1)=2$ and the reward rate $r(1, a_1) =-4$. Similarly,   $\lambda(1, a_2) = 7$ and $r(1, a_2) =-12$.  Here the rewards are negative representing costs. All the failure and repair times are assumed to be mutually independent. In addition, the transition probabilities for this problem are  given by $p(1 | 0 , a_i) = p(0|1, a_i)=1$ for $i=1, 2,$ i.e., the machine alternates between the operating mode and the repair mode. The length of the horizon for this CTMDP is set to be $H=1,$ and the objective of the CTMDP is to maximize the expected cumulative reward over the horizon $[0, H].$

The learning problem under consideration assumes that  the reward rate is known, but the failure and repair rates as well as the transition probabilities are unknown to the agent. We consider the episodic RL setting where every episode restarts with a fixed initial state $x_0=0.$ We implement the proposed CT-UCBVI algorithm with parameters $\delta =0.05, \epsilon_k = 1/{\sqrt{k}}, \lambda_{\max} = 7, \mathcal{S}=\{0, 1\}, \mathcal{A} = \{a_1, a_2\}$ and $H=1$. To fit Assumption~\ref{eq:assume}, we rescale the reward rates to $[0,1]$ by a linear transformation so that
$r(0, a_1) = 17/20,  r(0, a_2) = 1, r(1, a_1)= 8/20$ and $r(1, a_2) = 0$.

\begin{figure}[h]
    \begin{center}
    \includegraphics[width=0.7\textwidth, height=7cm]{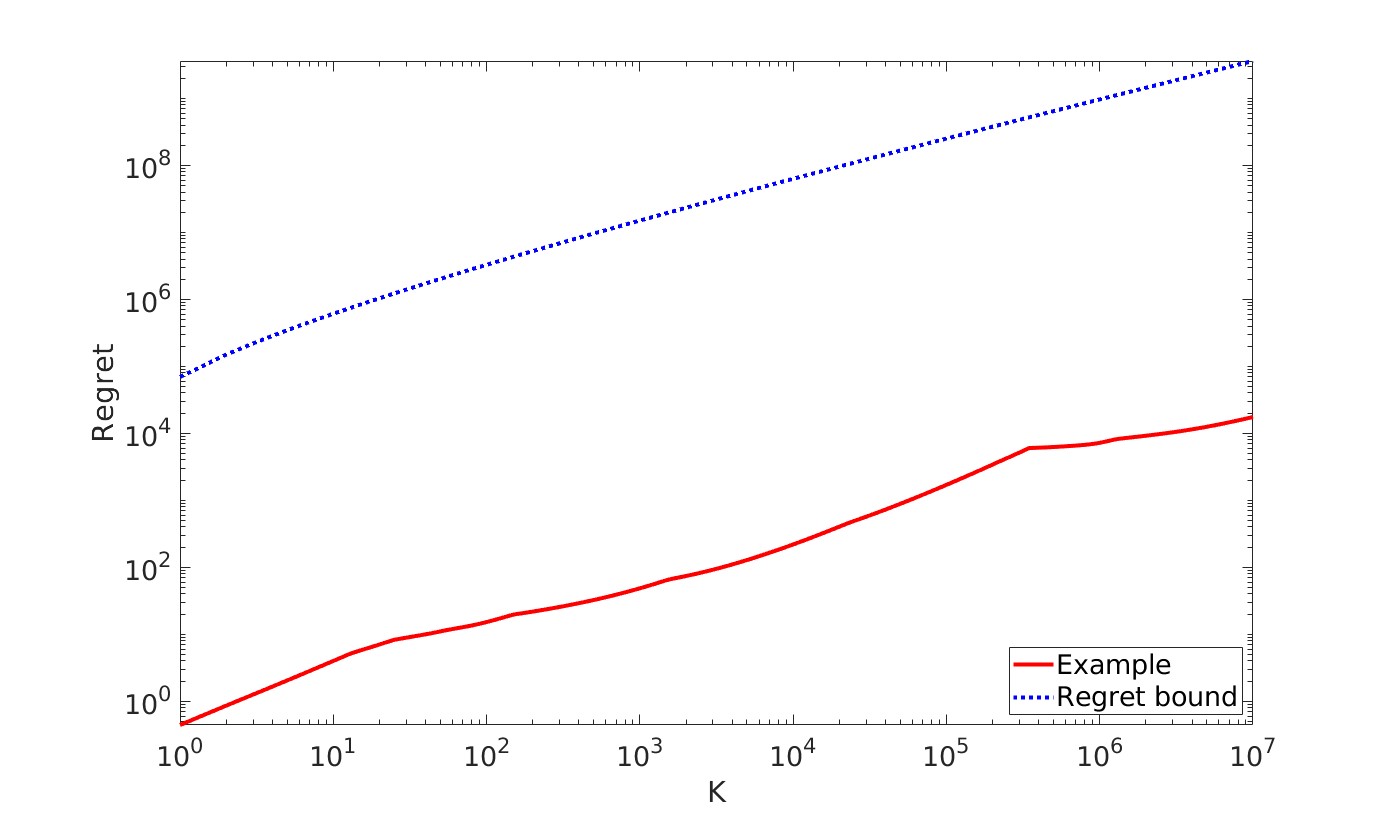}
    \caption{ Performance of the CT-UCBVI algorithm plotted on a log-log scale.  The line ``Example" represents the average regret of the CT-UCBVI algorithm on the machine replacement/repair example. The dash line ``Regret bound" plots the theoretical regret upper bound in \eqref{eq:regret-bound} as a function of  the number of episodes $K$. }
    \label{fig:1}
    \end{center}
\end{figure}

Figure~\ref{fig:1} illustrates the performance of the CT-UCBVI algorithm on this example. The expected regret is computed by averaging the regret values over 30 independent runs of the algorithm. We do not show the standard deviation
estimated from these 30 runs because it is very small and hardly visible  in the plot. For comparison, we also plot the regret upper bound in the right hand side of  \eqref{eq:regret-bound} as a function of the number of episodes $K$. Because the bound in \eqref{eq:regret-bound} is a worst-case bound whose magnitude is much greater than that of the expected regret of the algorithm on this specific instance, we plot the two curves on a log-log scale. 
As we can observe from Figure~\ref{fig:1}, the growth rate of the expected regret of the CT-UCBVI algorithm on this specific example is similar to that of the worst-case bound up to 10 million episodes. However, the exact rate in terms of $K$ requires a separate instance-dependent regret analysis of the algorithm and is left for future research.


\section{Proofs} \label{sec:proof-mainresult}

In this section we present proofs of the main results, Theorems~\ref{thm:CT-UCB} and \ref{thm:lowerbound}.

\subsection{Preliminary lemmas} \label{sec:pre-lemma}
In this subsection, we state and prove a few lemmas needed for the main proofs.

The first lemma provides a high probability bound on the estimator $\hat \lambda_k(x, a)$ in \eqref{eq:rate-est} for the holding-time rate.
Recall $L(\delta)= 4 \ln (2SAK/\delta)$, and  define
\begin{align} \label{eq:beta-k}
\beta_k(x,a,\delta)& := \sqrt{\frac{ \lambda_{\max} L(\delta) }{   \max\{T^k(x,a),  L(\delta)/ \lambda_{\max} \}}} =
\begin{cases}
\lambda_{\max} \quad & \text{if} \quad T^k(x,a) <  \frac{L(\delta)}{\lambda_{\max}}, \\
  \sqrt{ \frac{  \lambda_{\max} L(\delta) }{T^k(x,a)}}\quad & \text{otherwise}.
  \end{cases}
\end{align}
Then we have the following result.
\begin{lemma} \label{lem:rateUCB}
For any $\delta \in (0,1)$, we have
\begin{align} \label{eq:rateUCB}
\mathbb{P} \left( | \lambda(x, a) - \hat \lambda_k(x, a)| \le \beta_k(x,a,\delta), \quad \text{for all $x,a,k$} \right) \ge 1- \delta.
 \end{align}
\end{lemma}

 Note that the samples of holding time at a given state--action pair may not be i.i.d. due to the  truncation at time $H$. Hence, we are unable to directly apply concentration inequalities for i.i.d. exponential distributions to derive the confidence bound \eqref{eq:rateUCB}. The key idea to get around is to `piece together' the samples of exponential holding time at a pair $(x,a)$ as realizations of inter-arrival times of a Poisson process with rate $\lambda(x,a)$ modelling the number of visits to the pair over time. Even if the time is eventually truncated, the number of visits still follows a Poisson distribution. This enables us to apply the tail bounds for Poisson distribution, which is presented below for reader's convenience.

\begin{lemma}[Tail bounds for Poisson distribution \citealp{Canonne2017}] \label{lem:poisson}
Let the random variable $X$ follow a Poisson distribution  with mean $\lambda$. Then we have
\begin{align*}
\mathbb{P}(|X- \lambda| >x) \le e^{-\frac{x^2}{2(\lambda+x)}}, \quad x>0.
\end{align*}
\end{lemma}

\noindent\paragraph{Proof of Lemma~\ref{lem:rateUCB}.}
Fix $x,a, k,\delta$.
If $T^k(x,a)=0,$ i.e., the state--action pair $(x,a)$ has not been sampled before episode $k$, then clearly  $\hat \lambda_k(x,a)=0$ and  $\beta_k(x,a,\delta ) = \lambda_{\max}.$ Since $\lambda(x,a) \le \lambda_{\max}$ for all $(x,a)$, it follows trivially that $|\lambda(x,a) - \hat \lambda_k(x,a) | \le \beta_k(x,a,\delta )$.

So we focus on the non-trivial case when $T^k(x,a)>0.$ For each episode $i<k,$ we  piece together the holding times at $(x,a)$ and denote by $v_i(x,a)$ the total amount of time staying at $(x, a)$ in episode $i$. {Denote by $n_i(x,a)$ the total number of visits to $(x,a)$ in episode $i$, excluding the visit count to $(x,a)$ if the next state is not observed due to the finite horizon.
Given $v_i(x,a),$ the variable $n_i(x,a)$  follows a Poisson distribution with mean $\lambda(x,a) v_i(x,a)$. Since the holding times at $(x,a)$ across different episodes are independent, we obtain that given $(v_i(x,a): i <k)$, the counts $N^{k, +}(x,a) = \sum_{i <k} n_i(x,a)$,} follows a Poisson distribution with mean $\lambda (x,a) T^k(x,a)$ where $T^k(x,a) = \sum_{i<k} v_i(x,a)$. 



Given $T^k(s, a) =T>0,$ $\new{N^{k, +}(x,a)}$ follows a Poisson distribution with mean $\lambda(x,a) T. $ We now estimate the following conditional probability, noting  \eqref{eq:rate-est}:
\begin{align} \label{eq:rate-bound-1}
& \mathbb{P}(|\hat \lambda_k(x,a) - \lambda(x,a) | >\beta_k(x,a,\delta) | T^k(s, a) =T) \nonumber \\
& = \mathbb{P}(|\new{N^{k, +}(x,a)}/T^k(x,a)- \lambda(x,a)  | >\beta_k(x,a,\delta) ,\new{N^{k, +}(x,a)}/T^k(x,a) \le  \lambda_{\max} |  T^k(s, a) =T)  \nonumber \\
& \quad + \mathbb{P}(|\lambda_{\max}- \lambda(x,a)  | >\beta_k(x,a,\delta) , \new{N^{k, +}(x,a)} /T^k(x,a) \ge \lambda_{\max} |  T^k(s, a) =T)  \nonumber \\
& \le \mathbb{P}(|\new{N^{k, +}(x,a)}/T^k(x,a)- \lambda (x,a)  | >\beta_k(x,a,\delta) |  T^k(s, a) =T)  \nonumber \\
& \quad +  \mathbb{P}( \new{N^{k, +}(x,a)}/T^k(x,a) - \lambda (x,a)   \ge \lambda_{\max} - \lambda (x,a) ,  \lambda_{\max}- \lambda(x,a) > \beta_k(x,a,\delta) |  T^k(s, a) =T)  \nonumber \\
& \le 2 \mathbb{P}(| \new{N^{k, +}(x,a)}/T^k(x,a)- \lambda (x,a)  | >\beta_k(x,a,\delta) |  T^k(s, a) =T) .
\end{align}
Recall that  $\beta_k(x,a,\delta) = \sqrt{ \frac{  \lambda_{\max} L(\delta) }{T^k(x,a)}} <  \lambda_{\max}$ when $T^k(x,a) > \frac{L(\delta)}{ \lambda_{\max}}$.
Applying Lemma~\ref{lem:poisson}, we  obtain for $T > \frac{L(\delta)}{ \lambda_{\max}}$,
\begin{align*}
&  \mathbb{P}(|\new{N^{k, +}(x,a)}/T^k(x,a)- \lambda (x,a)  | >\beta_k(x,a,\delta) |  T^k(s, a) =T)\\
&  =  \mathbb{P}(|\new{N^{k, +}(x,a)} - \lambda (x,a) T | > \beta_k(x,a,\delta) T  |  T^k(s, a) =T)\\
& \le \exp \left(- \frac{ ( \beta_k(x,a,\delta) T)^2}{2 (\lambda (x,a) T +  \beta_k(x,a,\delta) T  ) }  \right) \\
& \le \exp \left(- \frac{ ( \frac{  \lambda_{\max} L(\delta) }{T}) T}{2 (\lambda (x,a) +  \lambda_{\max}  ) }  \right) \\
& \le e^{- L(\delta)/4} = \frac{\delta}{2 SA K},
\end{align*}
where we have used the facts that $\lambda(x,a) \le  \lambda_{\max}$ and $ \beta_k(x,a,\delta) \le \lambda_{\max}$ for all $(x,a)$ in the last two inequalities.
Together with \eqref{eq:rate-bound-1}, the above inequality  yields that for $T>\frac{L(\delta)}{ \lambda_{\max}},$
\begin{align} \label{eq:rate-bound-2}
\mathbb{P}(|\hat \lambda_k(x,a) - \lambda(x,a) | >\beta_k(x,a,\delta) | T^k(s, a) =T) \le \frac{\delta}{ SA K}.
\end{align}
This inequality actually also holds for  $T \le \frac{L(\delta)}{ \lambda_{\max}}.$ To see this, first note that in this case we have $\beta_k(x,a,\delta ) = \lambda_{\max}$ by the definition \eqref{eq:beta-k}.
Hence, it follows from  \eqref{eq:rate-est} that, with probability one, $|\hat \lambda_k(x,a) - \lambda(x,a)| \le \lambda_{\max}$ for all $(x,a) \in \mathcal{S} \times \mathcal{A}$.

Define the following two events
\begin{align*}
&E = \left\{ \left|{\hat \lambda_k(x,a)} -{\lambda(x,a)} \right| \le  \beta_k(x,a,\beta), \quad \text{for all $x,a, k$} \right\}, \\
&E_{k,x,a}=  \left\{ \left|{\hat \lambda_k(x,a)} -{\lambda(x,a)} \right| \le  \beta_k(x,a,\beta) \right\}.
\end{align*}
For fixed $k, x,a,\delta$, \eqref{eq:rate-bound-2} implies
\begin{align*}
&\mathbb{P}(E_{k,x,a}^c) = \mathbb{E}[\mathbb{P}(E_{k,x,a}^c |T^k(x,a) )]
\le  \frac{\delta}{ SA K}. 
\end{align*}
Applying a union bound, we deduce that $\mathbb{P} \left( E^c  \right)  \le \sum_{k, s,a} P(E_{k,s,a}^c) \le SAK \cdot  \frac{\delta}{ SA K}= \delta. $ The proof is complete. \qed


The next lemma, which is a known result,  gives  $L_1$ concentration bounds for the empirical transition probabilities. Write  
\begin{align}\label{eq:alpha-k}
\alpha_k(x,a,\delta): = \sqrt{\frac{2 [ S \ln 2 + \ln\left(SA H K^2 /\delta \right)]}{ \max\{1, \add{N^{k, +}(x,a)}\}} } .
\end{align}
\begin{lemma}[Lemma 17 in \citealp{Jaksch2010}] \label{lem:CI-p}
For any  $\delta \in (0,1)$, we have
\begin{align} \label{eq:CI-p}
\mathbb{P} \left( || p(\cdot|x, a) -  \hat p_k(\cdot|x, a)  ||_1 \le \alpha_k(x,a,\delta)  , \quad \text{for all $x,a, k$} \right) \ge  1 - \delta.
\end{align}
\end{lemma}

Define
\begin{align} \label{eq:G}
\mathcal{G} = \left\{ | \lambda(x, a) - \hat \lambda_k(x, a)| \le \beta_k(x,a,\delta) \; \text{and} \;   || p(\cdot|x, a) -  \hat p_k(\cdot|x, a)  ||_1 \le \alpha_k(x,a,\delta)  , \; \text{for all $x,a, k$} \right\},
\end{align}
which is the event that all the estimated parameters  lie in the confidence bounds.
Lemmas~\ref{lem:rateUCB} and \ref{lem:CI-p} yield  that
$\mathbb{P}(\mathcal{G}^c) \le 2 \delta$.
With the  concentration bounds of the estimated parameters, we next bound the error in applying the operator $\mathcal{T}^a$ defined in \eqref{eq:Qa} due to  model estimation errors.

{
\begin{lemma} \label{lem:error-model}
Fix $t \le H$ and a function $f(\cdot, \cdot): \mathcal{S} \times [0, H] \rightarrow [0, H]$ satisfying $0 \le f(x, t) \le t$ for all $x,t$ and $f(\cdot, 0)=0$.
Then, conditional on the occurrence of the event $\mathcal{G}$,  we have
\begin{align} \label{eq:model-error1}
&  \left| \mathbb{E}_{ \tau \sim \exp( \hat \lambda_k (x,a)), Y \sim \hat p_k(\cdot|x,a)} [ f (Y, (t-\tau)^+)] -   \mathbb{E}_{ \tau \sim \exp(\lambda(x,a)), Y \sim p(\cdot|x,a)} [ f (Y, (t-\tau)^+)] \right|  \nonumber \\
&= \left| \sum_{y \in \mathcal{S}} \int_0^t \hat \lambda_k(x,a) e^{- \hat \lambda_k(x,a) s}  f(y, t-s) \hat p_k(y|x,a) ds -   \sum_{y \in \mathcal{S}} \int_0^t\lambda(x,a) e^{-\lambda(x,a) s}  f(y, t-s)p(y|x,a) ds \right|  \nonumber \\
& \le \frac{t^2}{2} \beta_k(x,a,\delta) + t \alpha_k(x,a,\delta), 
\end{align}
for all $(x,a) \in \mathcal{S} \times \mathcal{A}$ and $k=1, \ldots, K$,
where $\beta_k$ and $\alpha_k$ are given in \eqref{eq:beta-k} and \eqref{eq:alpha-k} respectively.
\end{lemma}
}

\noindent\paragraph{Proof of Lemma~\ref{lem:error-model}.}
We estimate
\begin{align}
& \left| \sum_{y} \int_0^t\hat \lambda_k(x,a) e^{-\hat \lambda_k(x,a) s}  f(y, t-s) \hat p_k(y|x,a) ds -   \sum_{y} \int_0^t\lambda(x,a) e^{-\lambda(x,a) s}  f(y, t-s)p(y|x,a) ds \right|  \nonumber \\
& \le \left| \sum_{y} \int_0^t [\hat \lambda_k(x,a) e^{- \hat \lambda_k(x,a) s}  - \lambda(x,a) e^{-\lambda(x,a) s} ] f(y, t-s) \hat p_k(y|x,a) ds  \right|  \nonumber \\
& \quad + \left|\sum_{y} \int_0^t\lambda(x,a) e^{-\lambda(x,a) s}  f(y, t-s)[\hat p_k(y|x,a)- p(y|x,a)] ds \right|  \nonumber \\
& \le  \int_0^t   \left| \hat\lambda_k(x,a) e^{- \hat \lambda_k(x,a) s}  - \lambda(x,a) e^{-\lambda(x,a) s} \right| \cdot \add{(t-s)} ds     \nonumber \\
& \quad + \int_0^t\lambda(x,a) e^{-\lambda(x,a) s}  \left|\sum_{y} f(y, t-s)[\hat p_k(y|x,a)- p(y|x,a)] \right| ds ,  \label{eq:model-error}
\end{align}
where in the last inequality we have used the properties that $0 \le f(x, t) \le t$ for all $x, t$, and that $\sum_{y} \hat p_k(y|x,a) =1.$


We next bound the right hand side of \eqref{eq:model-error}. For the first term, note that for a fixed $s \in [0, t]$, the function $g(\lambda) := \lambda e^{-\lambda s}$ has the first-order derivative $g'(\lambda) =(1 -\lambda s) e^{-\lambda s}$, which  is easily seen to satisfy  $|g'(\lambda)| \le 1$ for $\lambda \ge 0$. Hence
$g$ is Lipschitz continuous with a Lipschitz constant one.
As a result,
{
\begin{align}\label{eq:error1}
&   \int_0^t   \left| \hat  \lambda_k(x,a) e^{-\hat\lambda_k(x,a) s}  - \lambda(x,a) e^{-\lambda(x,a) s} \right| \cdot (t-s) ds \\
& \le   \int_0^t (t-s)  \cdot  | \hat  \lambda_k(x,a) - \lambda(x,a)|  ds \nonumber \\
& =  \frac{t^2}{2} \cdot  | \hat  \lambda_k(x,a) - \lambda(x,a)|\le  \frac{t^2}{2} \cdot  \beta_k(x,a,\delta), \nonumber
\end{align}
}
conditional on the occurrence of $\mathcal{G}$.

It remains to bound the second term on the right hand side of \eqref{eq:model-error}. Because $0 \le f(x, t) \le t$ for all $x,t$, it follows from H\"{o}lder's inequality that
\begin{align} \label{eq:error2}
 \left|\sum_{y} f (y, t-s) [\hat p_k(y|x,a)- p(y|x,a)] \right| \le || f(\cdot, t-s)||_{\infty} \cdot ||\hat p_k(\cdot |x,a)- p(\cdot |x,a) ||_1\le \add{ (t-s)} \cdot  \alpha_k(x,a,\delta),
\end{align}
conditional on $\mathcal{G}$. 
Combining \eqref{eq:model-error}, \eqref{eq:error1} and \eqref{eq:error2} yields \eqref{eq:model-error1}. \qed


With Lemma~\ref{lem:error-model}, we can show that the computed value function $\hat V_{n_k}^k$ from Step 8 of the CT-UCBVI algorithm is a high-probability upper bound of the optimal value function $V^*$ up to some error. This is stated in the next result.
\begin{lemma}\label{lem:optimism}
Conditional on the occurrence of the event $\mathcal{G}$, we have
\begin{align}\label{eq:optimism}
V^*(x, t ) \le \hat V_{n_k}^k ( x, t) + \epsilon_k \cdot e^{\lambda_{\max} H}, \quad \text{ for all $ (x, t)\in \mathcal{S} \times [0, H]$, $k=1, \ldots, K$,}
\end{align}
where $\epsilon_k$, $k=1, \ldots, K$, are the accuracy parameters given in Algorithm~\ref{alg: CTUCB}. \end{lemma}

\noindent\paragraph{Proof of Lemma~\ref{lem:optimism}.}
From the value iteration, we have
\begin{align*}
V^*_{n+1}(x,t) &= \sup_{a \in \mathcal{A}} (\mathcal{T}^a V^*_n)(x,t) \\
&=  \sup_{a \in \mathcal{A}} \left\{ r(x,a ) + \mathbb{E}_{ \tau \sim \exp(\lambda(x,a)), Y \sim p(\cdot|x,a)} [ V^*_{n} (Y, (t-\tau)^+)]\right\},
\end{align*}
and $\lim_{n \rightarrow \infty} V_n^* = V^*.$ Meanwhile, in episode $k$, the CT-UCBVI algorithm applies the modified value iteration procedure to the CTMDP $(\mathcal{S}, \mathcal{A}, r(\cdot,\cdot) + b^k(\cdot,\cdot,\delta), \hat \lambda_k(\cdot,\cdot), \hat p_k(\cdot|\cdot,\cdot), H, \gao{x_0})$  with the accuracy parameter $\epsilon_k$, where
\begin{align}\label{eq:vi-k}
\hat V^k_{n+1} (x,t) &= \min\{ t, \sup_{a \in \mathcal{A}}  (\mathcal{T}^{a,k} \hat V^k_n)(x,t) \},
\end{align}
with the operator $\mathcal{T}^{a,k}$ given in \eqref{eq:Tak}.
Because $\mathcal{T}^{a,k}$ is a monotone nondecreasing mapping (see Remark 3.1 in \citealp{huang2011finite}), we obtain by induction that
\begin{align}\label{eq:mono}
H \ge \hat V^k_{n+1} (x,t) \ge  \hat V^k_n(x,t).
\end{align}
Hence, the sequence $ \hat V^k_n(x,t)$ converges as $n \rightarrow \infty$ and
we denote its limit by $\hat V^{k}_{\infty}(x,t):= \lim_{n \rightarrow \infty} \hat V^k_n(x,t)$ for  $(x,t) \in \mathcal{S} \times [0, H]$.

We first prove that $V^*(x, t ) \le \hat V^{k}_{\infty} ( x, t)$ for all $(x,t) \in \mathcal{S} \times [0, H]$. It suffices to show $V^*_{n}(x,t) \le \hat V^k_{n}(x,t) $ for all $n$, which we now prove by induction on $n$.
 At $n=0,$ we have $V^*_{0}(x,t) = \hat V^k_{0}(x,t) =0 $ for all $(x,t) \in \mathcal{S} \times [0, H]$.
Assuming $V^*_{n}(x,t) \le \hat V^k_{n}(x,t) $ for all $x,t$, we next prove that $V^*_{n+1}(x,t) \le \hat V^k_{n+1}(x,t) $ for all $x,t$.


Fixing an action $a,$ we compute
\begin{align}\label{eq:diff}
& (\mathcal{T}^{a,k } \hat V^k_n)(x,t) - (\mathcal{T}^a V^*_n)(x,t) \nonumber \\
&=  [b^k(x,a,\delta) + r(x,a)] \cdot \int_0^te^{- \hat \lambda_k(x,a) s} ds - r(x,a) \cdot \int_0^te^{-\lambda(x,a) s} ds \nonumber \\
&\quad + \mathbb{E}_{ \tau \sim \exp(\hat \lambda_k (x,a)), Y \sim \hat p_k(\cdot|x,a)} [\hat V^k_{n} (Y, (t-\tau)^+)] -   \mathbb{E}_{ \tau \sim \exp(\lambda(x,a)), Y \sim p(\cdot|x,a)} [ V^*_{n} (Y, (t-\tau)^+)]  \nonumber \\
&\ge   b^k(x,a,\delta) \cdot \int_0^te^{- \lambda_{\max} s} ds + r(x,a)  \left[ \int_0^te^{- \hat \lambda_k(x,a) s} ds -  \int_0^te^{-\lambda(x,a) s} ds\right] \nonumber \\
& \quad + \mathbb{E}_{ \tau \sim \exp(\hat \lambda_k (x,a)), Y \sim \hat p_k(\cdot|x,a)} [ V^*_{n} (Y, (t-\tau)^+)] -   \mathbb{E}_{ \tau \sim \exp(\lambda(x,a)), Y \sim p(\cdot|x,a)} [ V^*_{n} (Y, (t-\tau)^+)],
\end{align}
where the last inequality is due to the induction hypothesis and $\hat \lambda_k (x,a) \le \lambda_{\max}$.
Because the reward function is bounded in $[0,1]$, we can use a similar argument as in \eqref{eq:reward-disc} to deduce that, conditional on  $\mathcal{G}$,
\begin{align*}
\left| r(x,a)  \left[ \int_0^te^{- \hat \lambda_k(x,a) s} ds -  \int_0^te^{-\lambda(x,a) s} ds\right] \right| \le \frac{t^2}{2} \beta_k(x,a,\delta).
\end{align*}
Moreover, $0 \le V_n^*(x, t) \le V^*(x,t) \le t$ for all $(x, t)$ and $n$. Lemma~\ref{lem:error-model} then implies
\begin{align*}
&  \left| \mathbb{E}_{ \tau \sim \exp(\hat \lambda_k (x,a)), Y \sim \hat p_k(\cdot|x,a)} [ V^*_{n} (Y, (t-\tau)^+)] -   \mathbb{E}_{ \tau \sim \exp(\lambda(x,a)), Y \sim p(\cdot|x,a)} [ V^*_{n} (Y, (t-\tau)^+)] \right| \\
& \quad \le  \frac{t^2}{2} \beta_k(x,a,\delta) + t \alpha_k(x,a,\delta).
\end{align*}
It then follows from \eqref{eq:diff} that
\begin{align*}
(\mathcal{T}^{a,k} \hat V^k_n)(x,t) - (\mathcal{T}^a V^*_n)(x,t) \ge b^k(x,a,\delta) \cdot \int_0^t e^{- \lambda_{\max} s} ds -  t^2 \beta_k(x,a,\delta) - t \alpha_k(x,a,\delta).
\end{align*}
Therefore, $(\mathcal{T}^{a,k} \hat V^k_n)(x,t) - (\mathcal{T}^a V^*_n)(x,t) \ge 0$ for all $(x,t) \in \mathcal{S} \times [0, H]$ if the following holds:
\begin{align} \label{eq:b-sup}
b^k(x,a,\delta) \ge \sup_{ t \in [0, H]}  \frac{t^2\beta_k(x,a,\delta) + t \alpha_k(x,a,\delta)}{\int_0^te^{- \lambda_{\max} s} ds}.
\end{align}
One can easily verify that the function to be maximized  on the right hand side of \eqref{eq:b-sup} is increasing  in $t \ge 0$, and hence the maximum is attained at  $t=H$, i.e.,
{\small \begin{align*}
\sup_{ t \in [0, H]}  \frac{t^2 \beta_k(x,a,\delta) + t \alpha_k(x,a,\delta)}{\int_0^te^{- \lambda_{\max} s} ds}  = \frac{H^2 \beta_k(x,a,\delta) + H \alpha_k(x,a,\delta)}{\int_0^H e^{- \lambda_{\max} s} ds} = \frac{\lambda_{\max}}{1-e^{-\lambda_{\max} H}} \left[ H^2 \beta_k(x,a,\delta) + H \alpha_k(x,a,\delta) \right] .
\end{align*}}
Hence \eqref{eq:b-sup} indeed holds by the definition of $b^k(x,a,\delta)$ in \eqref{eq:bonus-k}. This proves  $V^*_{n+1}(x,t) \le \hat V^k_{n+1}(x,t) $ for all $x,t$, and thereby the desired inequality $V^*(x,t) \le \hat V^{k}_{\infty}(x,t) $ for all $x,t$. 

We are now ready to prove \eqref{eq:optimism}. We first show that in each episode $k,$ when the modified value iteration stops in $n_k$ iterations with $|| \hat V^{k}_{n_k}- \hat V^{k}_{n_k -1} ||_{\infty}\le \epsilon_k$, it returns $\hat V^{k}_{n_k}$ with
\begin{align}\label{eq:error-vi}
|| \hat V^{k}_{\infty}- \hat V^{k}_{n_k} ||_{\infty}  \le {\epsilon_k} \cdot e^{\lambda_{\max} H }.
\end{align}
Indeed, the definition of $\mathcal{T}^{a,k}$ in \eqref{eq:Tak} yields that  for any functions $u, v \in \mathbb{M},$
   \begin{align*}
   \left| (\mathcal{T}^{a,k}u )(x,t) -  (\mathcal{T}^{a,k}v )(x,t) \right|  & = \left|  \sum_{y} \int_0^t \hat \lambda_k (x,a) e^{-\hat \lambda_k(x,a) s}  \cdot (u-v) (y, t-s) \cdot \hat p_k(y|x,a) ds  \right|\\
   & \le ||u - v||_{\infty} \cdot \int_0^t \hat \lambda_k (x,a) e^{-\hat \lambda_k(x,a) s} ds,
    \end{align*}
where we  have used the fact that $ \sum_{y} \hat p_k(y|x,a)=1 $. Consequently,
   \begin{align*}
  || \mathcal{T}^{a,k}u -  \mathcal{T}^{a,k}v ||_{\infty}  & \le ||u - v||_{\infty} \cdot  \sup_{x \in \mathcal{S}}\int_0^H\hat \lambda_k (x,a) e^{-\hat \lambda_k(x,a) s} ds \\
  & \le ||u - v||_{\infty} \cdot (1 - e^{- \lambda_{\max} H}),
    \end{align*}
    noting that $ \hat \lambda_k(x,a) \le \lambda_{\max}$ for all $x,a.$ This implies that the operator $\mathcal{T}^{a,k}$ is a contraction mapping with a contraction coefficient $\rho:= 1- e^{- \lambda_{\max}H} \in (0,1)$. It follows that the mapping from $u$ to $\sup_{a \in \mathcal{A}}\mathcal{T}^{a,k} u$ is also a contraction with the same coefficient $\rho$; see, e.g., Theorem 2 in \cite{denardo1967contraction}.  Because the function $\min\{H, x\}$ is Lipschitz in $x \ge 0$ with the Lipschitz constant one, we infer from \eqref{eq:vi-k} that the mapping from $\hat V^{k}_{n}$ to $\hat V^{k}_{n+1}$ in the modified value iteration is a contraction with the coefficient $\rho$.   A standard argument,  see, e.g., the proof of Theorem 6.3.1 in \cite{puterman2014markov},  yields  that \eqref{eq:error-vi} holds. As we have proved $V^*(x,t) \le \hat V^{k}_{\infty}(x,t) $, it  follows that $V^*(x,t) \le   \hat V^{k}_{n_k} (x,t ) + {\epsilon_k} \cdot e^{\lambda_{\max} H } $ for all $x,t$.
    The proof is complete. \qed

In contrast to the discrete-time RL setting (see, e.g., Lemma 18 in \citealp{azar2017minimax}), we have an additional error term ${\epsilon_k} \cdot e^{\lambda_{\max} H }$ appearing in \eqref{eq:optimism}.
This is because the modified value iteration in Algorithm~\ref{alg:VI} for solving finite-horizon CTMDPs converges asymptotically with a linear rate $\rho := 1- e^{- \lambda_{\max} H}$ when $n \rightarrow \infty$. When this iteration is stopped at a finite number of iterations $n_k$, it naturally leads to some approximation error. In the above proof, by showing that the operator \eqref{eq:Tak} is a contraction mapping with the coefficient $\rho$, we control this error using similar arguments as in the analysis of value iteration for discrete-time infinite-horizon discounted MDPs \citep[Chapter 6.3]{puterman2014markov}.

In the next and final lemma of this subsection, we bound the per episode regret. For each episode $k,$
 denote by $\mathcal{H}_{k-1}$ the history  up to the end of episode $k-1$, i.e. $\mathcal{H}_{k-1}=\{ (x_n^l, a_n^l, \tau_n^l)_{n}: l=1, \ldots, k-1 \}$. Note that the policy $\pi^k$ specified in \eqref{eq:pi-k} depends on  $\mathcal{H}_{k-1}$. {Also recall the definitions of $\alpha_k, \beta_k$ and
$b^k$ in
\eqref{eq:alpha-k}, \eqref{eq:beta-k} and \eqref{eq:bonus-k} respectively. }

\begin{lemma} \label{lem:v-v-continuoustime}
Conditional on the occurrence of  $\mathcal{G}$,  for each episode $k,$ we have
{
\begin{align*}
 & V^{*}(x_0^k, H) - V^{\pi_k}(x_0^k, H)  \\
& \quad \le  \sum_{m=0}^{\infty} \mathbb{E}   \left( [b^k(x_m^k, a_m^k,\delta) + \alpha_k(x_m^k, a_m^k,\delta)  +  ( H - t_m^k)^+ \cdot \beta_k(x_m^k, a_m^k,\delta)   ] \cdot ( H - t_m^k)^+ \big|  \mathcal{H}_{k-1} \right)\\
&\qquad  +  \epsilon_k \cdot e^{\lambda_{\max} H},
\end{align*}
}
where $(x_i^k, a_i^k, \tau_i^k)_{i \ge 0}$ consists of  the trajectories  of the state, action and holding time corresponding to  the policy $\pi^k$, $t_m^k = \sum_{i=0}^{m-1} \tau_i^k $, and the expectation is taken over the randomness in the holding times $(\tau_i^k)_{i \ge 0}$ and state transitions $(x_i^k)_{i \ge 1}$. 
\end{lemma}

\noindent\paragraph{Proof of Lemma~\ref{lem:v-v-continuoustime}.}
Lemma~\ref{lem:optimism} yields $V^{*}(x, H)  \le \hat V_{n_k}^{k}(x, H) +  \epsilon_k \cdot e^{\lambda_{\max} H}$. So we only need to bound the quantity $\hat V_{n_k}^{k}(x_0^k, H) -V^{\pi_k}(x_0^k, H )$. 

Recall that for $i \le n_k-1$, we have the modified value iteration:
{
\begin{align*}
\hat V^k_{i+1} (x,t) &= \min\{ t, \sup_{a \in \mathcal{A}} (\mathcal{T}^{a,k} \hat V^k_i ) (x,t) \} \quad \text{and} \quad
\pi^k( x, t )   =  \arg \max_{a \in \mathcal{A}} \{ (\mathcal{T}^{a,k} \hat V_{n_k}^k) (x,t) \}, \quad \text{for all $x,t.$}
\end{align*}
By \eqref{eq:mono} and the monotonicity of the operator $\mathcal{T}^{a,k}$, we have
\begin{align}  \label{eq:rec-Vk}
 \hat V^{k}_{n_k}(x_0^k, H)  &\le  \sup_{a \in \mathcal{A}} (\mathcal{T}^{a,k} \hat V^k_{n_k-1})(x_0^k, H) \le  \sup_{a \in \mathcal{A}} (\mathcal{T}^{a,k} \hat V^k_{n_k})(x_0^k, H) \nonumber \\
 & = \left[  r(x_0^k, a_0^k) + b^k(x_0^k, a_0^k,\delta) \right]   \cdot  \int_0^H e^{-\hat \lambda_k(x_0^k, a_0^k) s} ds \nonumber \\
& \quad + \mathbb{E}_{ \tau_0^k \sim \exp(\hat \lambda_k (x_0^k, a_0^k)), x_1^k \sim \hat p_k(\cdot|x_0^k, a_0^k)} [ \hat V_{n_k}^k (x_1^k, (H-  \tau_0^k)^+)],
\end{align}
where $a_0^k := \pi^k(x_0^k, H)$ with $H$ being the remaining horizon at time 0. }
In addition,
given $\mathcal{H}_{k-1}$, the policy $\pi_k$ in \eqref{eq:pi-k} is Markovian and we have the following relation (see Lemma 3.1 in \citealp{huang2011finite}):
\begin{align}\label{eq:Vpik}
V^{\pi_k}(x_0^k, H) = r(x_0^k, a_0^k)   \cdot  \int_0^ H e^{-\lambda(x_0^k, a_0^k) s} ds + \mathbb{E}_{\tau_0^k \sim \exp(\lambda (x_0^k, a_0^k)), x_1^k \sim p(\cdot|x_0^k, a_0^k)} [ V^{\pi^k} (x_1^k, (H- \tau_0^k)^+)].
\end{align}
In view of \eqref{eq:rec-Vk} and \eqref{eq:Vpik}, we can directly compute that
\begin{align}\label{eq:reward-disc}
 &\left[  r(x_0^k, a_0^k) + b^k(x_0^k, a_0^k,\delta) \right]   \cdot  \int_0^H e^{-\hat \lambda_k(x_0^k, a_0^k) s} ds - r(x_0^k, a_0^k)   \cdot  \int_0^H e^{-\lambda(x_0^k, a_0^k) s} ds \nonumber \\
 & = r(x_0^k, a_0^k)  \cdot  \left|\int_0^H e^{-\hat \lambda_k(x_0^k, a_0^k) s} ds - \int_0^H e^{- \lambda (x_0^k, a_0^k) s} ds \right| + b^k(x_0^k, a_0^k,\delta) \cdot   \int_0^H e^{-\hat \lambda_k(x_0^k, a_0^k) s} ds \nonumber \\
  & \le r(x_0^k, a_0^k)  \cdot  \frac{H^2}{2} \left| \hat \lambda_k(x_0^k, a_0^k)  - \lambda (x,a)  \right| + b^k(x_0^k, a_0^k,\delta ) \cdot H \nonumber \\
  & \le  \frac{H^2}{2} \beta_k(x_0^k, a_0^k,\delta)+ b^k(x_0^k, a_0^k,\delta) \cdot H,
 \end{align}
where in the second inequality we use the fact that the function $\int_0^{H} e^{-\lambda s } ds$ is Lipschitz continuous in $\lambda \in [0, \lambda_{\max}]$ with a Lipschitz constant $H^2/2$, and in the third inequality we use the assumption that the event $\mathcal{G}$ occurs and the fact that rewards are bounded by one. 
Then we deduce from \eqref{eq:rec-Vk} and \eqref{eq:Vpik} that,  given $\mathcal{H}_{k-1}$,
{
\begin{align*}
&\hat V_{n_k}^{k}(x_0^k, H ) -V^{\pi_k}(x_0^k, H) \\
&\le  \frac{H^2}{2} \beta_k(x_0^k, a_0^k,\delta)+ b^k(x_0^k, a_0^k,\delta) \cdot H \\
& \quad + \mathbb{E}_{ \tau_0^k \sim \exp(\hat \lambda_k (x_0^k, a_0^k)), x_1^k \sim \hat p_k(\cdot|x_0^k, a_0^k )} [ \hat V_{n_k}^k (x_1^k, (H- \tau_0^k)^+)] \\
& \quad  -  \mathbb{E}_{\tau_0^k \sim \exp(\lambda (x_0^k, a_0^k)), x_1^k \sim p(\cdot|x_0^k, a_0^k )} [ V^{\pi^k} (x_1^k, (H- \tau_0^k)^+)]\\
&=   \frac{H^2}{2} \beta_k(x_0^k, a_0^k,\delta)+ b^k(x_0^k, a_0^k,\delta) \cdot H  \\
& \quad + \mathbb{E}_{\tau_0^k \sim \exp(  \hat \lambda_k (x_0^k, a_0^k )),  x_1^k\sim \hat p_k(\cdot|x_0^k, a_0^k)} [ \hat V_{n_k}^k (x_1^k, (H- \tau_0^k)^+)]  \\
& \quad - \mathbb{E}_{\tau_0^k \sim \exp(\lambda (x_0^k, a_0^k)), x_1^k \sim p(\cdot|x_0^k, a_0^k )} [ \hat V_{n_k}^k (x_1^k, (H- \tau_0^k)^+)] \\
& \quad +   \mathbb{E}_{\tau_0^k \sim \exp(\lambda (x_0^k, a_0^k)), x_1^k \sim p(\cdot|x_0^k, a_0^k)} [ \hat V_{n_k}^k (x_1^k, (H- \tau_0^k)^+) -  V^{\pi_k} (x_1^k, (H- \tau_0^k)^+ )] \\
&\le  {H^2} \beta_k(x_0^k, a_0^k,\delta)+ H \cdot b^k(x_0^k, a_0^k,\delta)  +  H \cdot \alpha_k(x_0^k, a_0^k,\delta)  \\
& \quad  +   \mathbb{E}_{\tau_0^k \sim \exp(\lambda (x_0^k, a_0^k)), x_1^k \sim p(\cdot|x_0^k, a_0^k)} [ \hat V_{n_k}^k (x_1^k, (H- t_1^k)^+) -  V^{\pi_k} (x_1^k,(H- t_1^k)^+ )],
\end{align*}
where the last inequality follows from Lemma~\ref{lem:error-model} and the fact that $t_1^k = \tau_0^k$.
Using a similar argument, we can bound $\hat V_{n_k}^k (x_1^k, (H- t_1^k)^+) -  V^{\pi_k} (x_1^k,(H- t_1^k)^+ )$:
\begin{align*}
&\hat V_{n_k}^k (x_1^k, (H- t_1^k)^+) -  V^{\pi_k} (x_1^k,(H- t_1^k)^+ ) \\
&\le  { ((H- t_1^k)^+)^2} \beta_k(x_1^k, a_1^k,\delta)+ (H- t_1^k)^+ \cdot b^k(x_1^k, a_1^k,\delta)  +  (H- t_1^k)^+ \cdot \alpha_k(x_1^k, a_1^k,\delta)  \\
& \quad  +   \mathbb{E}_{\tau_1^k \sim \exp(\lambda (x_1^k, a_1^k)), x_2^k \sim p(\cdot|x_1^k, a_1^k)} [ \hat V_{n_k}^k (x_2^k, (H- t_2^k)^+) -  V^{\pi_k} (x_2^k,(H- t_2^k)^+ )],
\end{align*}
where $t_2^k = t_1^k + \tau_1^k = \tau_0^k + \tau_1^k.$}

{
Indeed,
we can apply the above procedure recursively to obtain, noting that {$\hat V_{n_k}^k(x, 0) = V^{\pi_k} (x,0) = 0$} for all $x \in \mathcal{S}$,  
\begin{align*} 
\hat V_{n_k}^{k}(x_0^k, H ) -V^{\pi_k}(x_0^k, H) 
&\le \sum_{m=0}^{\infty} \mathbb{E}\left[ b^k(x_m^k, a_m^k,\delta) \cdot  ( H - t_m^k)^+ \big|  \mathcal{H}_{k-1}  \right] \nonumber \\
& \quad +  \sum_{m=0}^{\infty} \mathbb{E}\left[ \beta_k(x_m^k, a_m^k,\delta) \cdot  [( H - t_m^k)^+]^2 \big|  \mathcal{H}_{k-1}  \right] \nonumber \\
& \quad +   \sum_{m=0}^{\infty} \mathbb{E}\left[ \alpha_k(x_m^k, a_m^k,\delta) \cdot  ( H - t_m^k)^+ \big|  \mathcal{H}_{k-1}  \right] ,
\end{align*} }
where the expectation is taken over the randomness in the  holding times $(\tau_i^k)_{i \ge 0}$ and state transitions $(x_i^k)_{i \ge 1}$ under $\pi_k$, conditional on $\mathcal{H}_{k-1}$.
The proof  is complete. \qed


\subsection{Proof of Theorem~\ref{thm:CT-UCB}}

{In this section, we prove Theorem~\ref{thm:CT-UCB}, with the objective being to bound the expected total regret over $K$ episodes. The main difficulty in the analysis is to bound the regret from estimation errors of transition probabilities and holding time rates for the CTMDP. In particular, the standard pigeon-hole argument can not be directly applied to our setting.
 To illustrate this point as well as both the subtlety and complexity in bounding the regret from estimation errors of transition probabilities of CTMDPs, we consider one sample path where $t_1^k >H$ for all $k=1, \ldots, K$ (recall that $t_1^k$ is the first transition time in episode $k$, and it follows an exponential distribution). In such a case, we can infer from \eqref{eq:Nkplus} that  $N^{k,+}(x,a) =0$ for all $(x,a)$ and $k$, which implies that for this path,
\begin{align} \label{eq:sumK}
\sum_{k= 1}^{K}   \frac{1}{ \sqrt{\max\{1, N^{k,+}(x^k_{0}, a^k_{0} ) \}}} = K.
\end{align}
Hence, the standard pathwise pigeon-hole argument used for obtaining {\it sublinear} regret in episodic DTMDPs (see, e.g., \cite{azar2017minimax}) no longer works  in our setting. We overcome this difficulty by devising a new probabilistic argument to bound the regret from estimation errors of transition probabilities (which needs to be weighted by the remaining time horizon $( H- t_{m}^k)^+$). The result is presented as Proposition \ref{prop:proberror}.
}

{
\begin{proposition}\label{prop:proberror}
\begin{align*} 
 \mathbb{E} \left[  \sum_{k= 1}^{K} \sum_{m=0}^{\infty}  \frac{ ( H- t_{m}^k)^+ }{  \sqrt{\max\{1, N^{k,+}(x^k_{m}, a^k_{m} ) \}}}   \right]
 \le  3 \left( H+  \frac{1}{\lambda_{\min} }\right)  \cdot \sqrt{SA K} \cdot    \frac{ (\lambda_{\max} H +1)  \ln K}{ \ln(\ln(K) +1)} .
\end{align*}
\end{proposition}
}

\vspace{2mm}
{
To bound the regret from estimation error in the holding time rate, which involves the continuous varible $T^k(x,a)$ (the cumulative amount of time spent at $(x,a)$),  we use an idea similar to that in the proof of Lemma~\ref{lem:rateUCB}. That is, we convert bounding the function of $T^k(x,a)$ to bounding the function of $N^{k, +}(x,a)$ by exploiting  properties of Poisson distributions. The result is given in Proposition~\ref{prop:rateerror}.
}

{
\begin{proposition} \label{prop:rateerror}
\begin{align*} 
& \mathbb{E}   \left[ \sum_{k= 1}^{K} \sum_{m=0}^{\infty}   \frac{(H - t_m^k)^+}{\sqrt{  \max\{T^k(x_m^k,a_m^k),  L(\delta)/ \lambda_{\max} \}} } \right]  \\
& \quad \le \sqrt{ 2 \lambda_{\max}}  \cdot  3 \left( H+  \frac{1}{\lambda_{\min} }\right)  \cdot \sqrt{SA K} \cdot   \frac{ (\lambda_{\max} H +1)  \ln K}{ \ln(\ln(K) +1)} .
\end{align*}
\end{proposition}
}

{
Proofs of Propositions~\ref{prop:proberror} and \ref{prop:rateerror} are deferred to the end of this section. \\
}

\noindent\paragraph{Proof of Theorem~\ref{thm:CT-UCB}.}
Write the total regret as
\begin{align} \label{eq:r1}
 & \text{Regret}(\mathcal{M}, \text{CT-UCBVI}, K)\nonumber  \\
 &= \mathbb{E} \left[ \sum_{k=1}^K V^{*}(x_0^k, H) -  \sum_{k= 1}^K  V^{\pi_k}(x_0^k, H)   \right] \nonumber \\
  & = \mathbb{E} \left[ 1_{\mathcal{G}}  \cdot  \sum_{k= 1}^{K} (V^{*}(x_0^k, H) - V^{\pi_k}(x_0^k, H)   )  \right]  +  \mathbb{E} \left[ 1_{\mathcal{G}^c
 }  \cdot \sum_{k= 1}^{K} (V^{*}(x_0^k, H) - V^{\pi_k}(x_0^k, H))   \right]  \nonumber \\
& \le   \mathbb{E} \left[ 1_{\mathcal{G}}  \cdot  \sum_{k= 1}^{K} (V^{*}(x_0^k, H) - V^{\pi_k}(x_0^k, H)   )    \right]  +  2 \delta K H,
\end{align}
where the inequality holds because $\mathbb{P}(\mathcal{G}^c) \le 2 \delta$ and  $0 \le V^{\pi^k} (x_0^k, H) \le V^*(x_0^k, H ) \le H$ (by the assumption that the reward rates are bounded between zero and one).  In the following we  choose $\delta = 1/(2KH)$ so that $2 \delta K H =1$ in \eqref{eq:r1} and $L(\delta)= 4 \ln (2SAK/\delta) >1.$

{For notational simplicity, let $C:=\frac{\lambda_{\max}}{1-e^{-\lambda_{\max} H}} \vee 1$.
It follows from  Lemma~\ref{lem:v-v-continuoustime} and the definitions of $\alpha_k, \beta_k$ and
$b^k$ in \eqref{eq:alpha-k}, \eqref{eq:beta-k} and \eqref{eq:bonus-k} that,  conditional on  $\mathcal{G}$,
\begin{align*}
& V^{*}(x_0^k, H) - V^{\pi_k}(x_0^k, H) \\
& \le  {\epsilon_k} \cdot e^{\lambda_{\max} H } +
 (CH+1)\cdot  \sum_{m=0}^{\infty} \mathbb{E}\left[  \sqrt{\frac{2 [ S \ln 2 + \ln\left(SA H K^2 /\delta \right)]}{ \max\{1, { N^{k, +} }(x_m^k,a_m^k)\}} }  \cdot (H -t_m^k)^+ \Bigg| \mathcal{H}_{k-1} \right] \\
  &  \quad +  (CH^2+H)\cdot  \sum_{m=0}^{\infty} \mathbb{E}\left[  \sqrt{\frac{ \lambda_{\max} L(\delta) }{   \max\{T^k(x_m^k,a_m^k),  L(\delta)/ \lambda_{\max} \}}} \cdot (H - t_m^k)^+ \Bigg| \mathcal{H}_{k-1}  \right]
\end{align*}
where $T^k(x,a)$ and $N^{k, +}(x,a)$ are given in \eqref{eq:Timek} and \eqref{eq:Nkplus} respectively.
This implies
\begin{align} \label{eq:r2}
 &    \mathbb{E} \left[ 1_{\mathcal{G}}  \cdot  \sum_{k= 1}^{K} (V^{*}(x_0^k, H) - V^{\pi_k}(x_0^k, H)   )  \right] \nonumber \\
 & \le   \sum_{k= 1}^{K} {\epsilon_k} \cdot e^{\lambda_{\max} H } +
(CH+1)\cdot  \sum_{k= 1}^{K} \sum_{m=0}^{\infty} \mathbb{E}\left[  \sqrt{\frac{2 [ S \ln 2 + \ln\left(SA H K^2 /\delta \right)]}{ \max\{1, { N^{k, +} }(x_m^k,a_m^k)\}} }  \cdot (H -t_m^k)^+ \right]  \nonumber \\
  &  \quad +  (CH^2+H)\cdot  \sum_{k= 1}^{K} \sum_{m=0}^{\infty} \mathbb{E}\left[  \sqrt{\frac{ \lambda_{\max} L(\delta) }{   \max\{T^k(x_m^k,a_m^k),  L(\delta)/ \lambda_{\max} \}}} \cdot (H - t_m^k)^+  \right] .
\end{align}
Applying Propositions~\ref{prop:proberror} and ~\ref{prop:rateerror}, we can readily infer from \eqref{eq:r2} that
{\begin{align*}
& \text{Regret}(\mathcal{M}, \text{CT-UCBVI}, K) \\
& \le 3 \left( CH+1\right) \cdot \sqrt{SA K} \cdot  \left( H+  \frac{1}{\lambda_{\min} }\right) \cdot  (\lambda_{\max} H +1)  \frac{\ln K}{ \ln(\ln(K) +1)} . \nonumber \\
& \quad   \cdot \left(   \sqrt{ 2 S + 6\ln (2SA K H) }  + 4 \lambda_{\max} H  \sqrt{ \ln(2SAKH)} \right)  \\
& \quad \quad  + \sum_{k= 1}^{K} {\epsilon_k} \cdot e^{\lambda_{\max} H }+ 1.
\end{align*} }
The proof is  complete. } \qed

\subsubsection{Proofs of Propositions~\ref{prop:proberror} and \ref{prop:rateerror}}

\noindent\paragraph{Proof of Proposition~\ref{prop:proberror}.}
{Noting $( H- t_{m}^k)^+ = ( H- t_{m}^k) 1_{t_{m}^k < H} $ and $\lim_{x \rightarrow 0+} \frac{x}{1- e^{-\lambda_{\min} x}} = 1/ \lambda_{\min} \in (0, \infty),$
 we have
\begin{align} \label{eq:CS1}
  \frac{ (H - t_m^k) ^+ } {\sqrt{ \max\{1, N^{k,+}(x^k_{m}, a^k_{m} ) \}}}
 =  \frac{ (1- e^{-\lambda_{\min} (H - t_m^k)^+}) \cdot 1_{t_{m}^k < H} } {\sqrt{ \max\{1, N^{k,+}(x^k_{m}, a^k_{m} ) \}}} \cdot  \frac{(H - t_m^k)^+ }{1- e^{-\lambda_{\min} (H - t_m^k)^+}} .
\end{align}
Using the fact that $e^{\lambda_{\min} x} \ge 1 + \lambda_{\min} x$ for any $x \ge 0, $ one can easily verify that $x/(1- e^{-\lambda_{\min} x} ) \le x + \frac{1}{\lambda_{\min} }$ for any $x \ge 0$.  It follows that 
\begin{align*}
   \frac{(H - t_m^k)^+}{1- e^{-\lambda_{\min} (H - t_m^k)^+}} \le  (H - t_m^k)^+ +  \frac{1}{\lambda_{\min} } \le H+  \frac{1}{\lambda_{\min} }.
\end{align*}
This implies
\begin{align} \label{eq:CS1}
  \frac{ (H - t_m^k) ^+ } {\sqrt{ \max\{1, N^{k,+}(x^k_{m}, a^k_{m} ) \}}}
\le \frac{ (1- e^{-\lambda_{\min} (H - t_m^k)^+}) \cdot 1_{t_{m}^k < H} } {\sqrt{ \max\{1, N^{k,+}(x^k_{m}, a^k_{m} ) \}}} \cdot  \left(  H+  \frac{1}{\lambda_{\min} } \right).
\end{align}
 Because the holding time $t_{m+1}^k - t_m^k $ has an exponential distribution with rate parameter $\lambda(x_m^k, a_m^k),$ we can compute
\begin{align*}
& \mathbb{E} \left[   \frac{1_{t_{m +1}^k \le H}} { \sqrt{ \max\{1, N^{k,+}(x^k_{m}, a^k_{m} ) \}}} \Big| (x_m^k, a_m^k, t_m^k), N^{k,+}(x^k_{m}, a^k_{m} )  \right] \\
& =  \mathbb{P} ( t_{m+1}^k - t_m^k \le (H - t_m^k)^+ | x_m^k, a_m^k, t_m^k  ) \cdot \frac{1_{t_{m}^k < H} } { \sqrt{ \max\{1, N^{k,+}(x^k_{m}, a^k_{m} ) \}}}\\
& =   (1- e^{-\lambda(x_m^k, a_m^k) (H - t_m^k)^+}) \cdot \frac{1_{t_{m}^k  < H}} { \sqrt{ \max\{1, N^{k,+}(x^k_{m}, a^k_{m} ) \}}}\\
& \ge  \frac{  (1- e^{- \lambda_{\min} (H - t_m^k)^+}) \cdot 1_{t_{m}^k  < H}} {\sqrt{ \max\{1, N^{k,+}(x^k_{m}, a^k_{m} ) \}}},
\end{align*}
where the last inequality is due to the fact that $\lambda(x,a) \ge  \lambda_{\min}$ for all $(x,a)$. It then follows from \eqref{eq:CS1} that
\begin{align} \label{eq:CS2}
  \mathbb{E} \left[  \sum_{k= 1}^{K} \sum_{m=0}^{\infty}  \frac{ ( H- t_{m}^k)^+ }{  \sqrt{\max\{1, N^{k,+}(x^k_{m}, a^k_{m} ) \}}}   \right] \le
 \left( H+  \frac{1}{\lambda_{\min} }\right)  \cdot \mathbb{E} \left[   \sum_{m=0}^{\infty} \sum_{k= 1}^{K}   \frac{1_{t_{m +1}^k \le H}} { \sqrt{ \max\{1, N^{k,+}(x^k_{m}, a^k_{m} ) \}}}  \right] .
\end{align}
}
{
We next bound the right-hand side of \eqref{eq:CS2}. For a state--action pair $(x,a) \in \mathcal{S} \times \mathcal{A}$, denote by $Y_m^l (x,a) : = 1_{ \{( x_m^l, a_m ^l) = (x,a)\}} \cdot 1_{t_{m+1}^l \le H}$, and let $$N_m^{k,+}(x, a ) := \sum_{l=1}^{k-1} Y_m^l (x,a) = \sum_{l=1}^{k-1} 1_{ \{( x_m^l, a_m ^l) = (x,a)\}} \cdot 1_{t_{m+1}^l \le H}.$$  It is clear that $0 \le Y_m^k (x,a)  \le 1 \le \max\{1, N_m^{k,+}(x, a ) \} $ and  $N_m^{k, +}(x,a)  \le N^{k, +}(x,a) . $
We can then obtain
\begin{align*}
 \sum_{k= 1}^{K} \frac{1_{t_{m+1}^k \le H}} { \sqrt{{ \max\{1, N^{k,+}(x^k_{m}, a^k_{m} ) \}}} } & \le  \sum_{k= 1}^{K} \frac{1_{t_{m+1}^k \le H}} {\sqrt{{ \max\{1, N_m^{k,+}(x^k_{m}, a^k_{m} ) \}}} }  \\
& \le  \sum_{ (x,a)}  \sum_{k= 1}^{K} \frac{Y_m^k (x,a)} {\sqrt{ \max\{1, N_m^{k,+}(x, a ) \}}} \\
& \le 3  \sum_{ (x,a)}  \sqrt{  N_m^{(K+1),+}(x, a )} \\
& \le 3 \sqrt{SA \cdot  \sum_{ (x,a)}   N_m^{(K+1),+}(x, a ) }  \\
& =3 \sqrt{SA \cdot  \sum_{l=1}^{K} 1_{t_{m+1}^l \le H}  } ,
\end{align*}
where the third inequality is due to Lemma 19 in \cite{Jaksch2010}, and the fourth inequality follows from Cauchy--Schwarz inequality. Thus, we have
\begin{align} \label{eq:CS3}
\mathbb{E} \left[   \sum_{m=0}^{\infty} \sum_{k= 1}^{K}   \frac{1_{t_{m +1}^k \le H}} { \sqrt{ \max\{1, N^{k,+}(x^k_{m}, a^k_{m} ) \}}}  \right] \le  3 \sqrt{SA} \cdot  \mathbb{E} \left[   \sum_{m=0}^{\infty} \sqrt{   \sum_{l=1}^{K} 1_{t_{m+1}^l \le H} } \right].
\end{align}
Given the policy $\pi^k,$ denote by $J^k(H): = \sup\{m: t_m^k \le H\}$, and set $J^K_{M}(H): = \max\{ J^k(H): k =1, 2, \ldots, K\}$.  It is clear that
\begin{align} \label{eq:JM-eq}
 \sum_{m=0}^{\infty} \sqrt{   \sum_{l=1}^{K} 1_{t_{m+1}^l \le H} } =  \sum_{m=0}^{J^K_{M}(H) - 1 } \sqrt{   \sum_{l=1}^{K} 1_{t_{m+1}^l \le H} }   \le \sqrt{K} \cdot J^K_{M}(H) .
\end{align}
In the equation above, we have allowed the agent to make (the same) $(J_M^K(H)+1)$ number of decisions for each episode, and hence it is possible that some of the decision steps are taken after the time $H$. Such decision steps are {\it fictitious}, introduced only for the purpose of a cleaner  analysis.
We then obtain from \eqref{eq:CS3} that
\begin{align} \label{eq:CS4}
\mathbb{E} \left[   \sum_{m=0}^{\infty} \sum_{k= 1}^{K}   \frac{1_{t_{m +1}^k \le H}} { \sqrt{ \max\{1, N^{k,+}(x^k_{m}, a^k_{m} ) \}}}  \right] \le  3 \sqrt{SA K} \cdot  \mathbb{E}( J^K_{M}(H) ).
\end{align}
It remains to bound $\mathbb{E}( J^K_{M}(H)).$
Noting  $\lambda(x,a) \le \lambda_{\max}$ for all $(x,a)$,  we get  that $J^k(H) \le Z^k(H)$ almost surely for each $k=1, \ldots, K$, where $\{Z^k(\cdot): k =1, \ldots, K\}$ are $K$ independent Poisson processes with a common rate $ \lambda_{\max}$. It then follows from the definition of $J^K_M(H)$ that
\begin{align} \label{eq:term1-3}
\mathbb{E}[ J^K_M(H) ] \le \mathbb{E}\left[ \max_{k=1, \ldots, K}\{ Z^k(H) \} \right] \le \frac{ (\lambda_{\max} H +1) \ln K}{ \ln(\ln(K) +1)},
\end{align}
where the second inequality is a standard bound on the expected maximum value of several independent Poisson random variables with a common mean $\lambda_{\max} H$;  see e.g. Exercise 2.18 in \cite{boucheron2013concentration}. Therefore, we can infer from \eqref{eq:CS4} and \eqref{eq:CS2} that
\begin{align*}
  \mathbb{E} \left[  \sum_{k= 1}^{K} \sum_{m=0}^{\infty}  \frac{ ( H- t_{m}^k)^+ }{  \sqrt{\max\{1, N^{k,+}(x^k_{m}, a^k_{m} ) \}}}   \right] \le
3 \left( H+  \frac{1}{\lambda_{\min} }\right)  \cdot \sqrt{SA K} \cdot     \frac{ (\lambda_{\max} H +1) \ln K}{ \ln(\ln(K) +1)} .
\end{align*}
The proof is complete. } \qed


\noindent\paragraph{Proof of Proposition~\ref{prop:rateerror}.}

{ 
Fix $(x,a) \in \mathcal{S} \times \mathcal{A}$ and $k.$ As already shown  in the proof of Lemma~\ref{lem:rateUCB}, given $T^k(x,a)$, $N^{k, +}(x,a)$ follows a Poisson distribution with mean $\lambda (x,a) T^k(x,a)$.
Hence we have
\begin{align*}
 \mathbb{E}   \left[ \frac{1}{\sqrt{\max\{ N^{k, +}(x,a), 1 \} }}  \Bigg|T^k(x,a) \right]
 & \ge \frac{1}{\sqrt{ \mathbb{E}   \left[ \max\{ N^{k, +}(x,a), 1 \}  | T^k(x,a)\right]  }  } \\
& = \frac{1}{\sqrt{ e^{- \lambda(x,a) T^k(x,a) }  + \lambda(x,a) T^k(x,a)  }} \\
& \ge \frac{1}{\sqrt{ 2 \max\{ \lambda(x,a) T^k(x,a) , 1\} }} \\
& \ge \frac{1}{\sqrt{ 2 \lambda_{\max}}} \cdot \frac{1}{\sqrt{\max\{ T^k(x,a), 1/\lambda_{\max} \} }} ,
\end{align*}
where the first inequality is due to Jensen's inequality, the second inequality follows from the fact that $e^{-t} + t \le 2 \max \{ t, 1\}$ for $t \ge 0$, and the last inequality is due to $\lambda(x,a) \le \lambda_{\max}.$ It follows that
\begin{align}\label{eq:cond-exp}
& \mathbb{E}   \left[   \frac{(H - t_m^k)^+  }{ \sqrt{ \max\{N^{k, +}(x_m^k,a_m^k),  1\}} } \right] \nonumber  \\
& = \mathbb{E}   \left( \mathbb{E}   \left[   \frac{ (H - t_m^k)^+  }{ \sqrt{ \max\{ N^{k, +}(x_m^k,a_m^k),  1\}} } \Big| (x_m^k,a_m^k, t_m^k), T^k(x_m^k,a_m^k)\right]  \right)  \nonumber  \\
& \ge  \frac{1}{\sqrt{ 2 \lambda_{\max}}} \cdot  \mathbb{E}   \left(  \frac{ (H - t_m^k)^+  }{  \sqrt{  \max\{T^k(x_m^k,a_m^k),  1/ \lambda_{\max} \}} }  \right) .
\end{align}
Consequently,
\begin{align*}
& \sum_{k= 1}^{K} \sum_{m=0}^{\infty}  \mathbb{E}   \left[  \frac{(H - t_m^k)^+  }{  \sqrt{ \max\{T^k(x_m^k,a_m^k),  L(\delta)/ \lambda_{\max} \}} }  \right] \nonumber \\
& \le \sum_{k= 1}^{K} \sum_{m=0}^{\infty}  \mathbb{E}   \left[  \frac{ (H - t_m^k)^+ }{  \sqrt{ \max\{T^k(x_m^k,a_m^k),  1/ \lambda_{\max} \}} }  \right]  \nonumber \\
& \le \sum_{k= 1}^{K} \sum_{m=0}^{\infty}  \sqrt{ 2 \lambda_{\max}} \cdot \mathbb{E}   \left[   \frac{(H - t_m^k)^+}{ \sqrt{ \max\{N^{k, +}(x_m^k,a_m^k),  1\}} }   \right] \nonumber  \\
& \le \sqrt{ 2 \lambda_{\max}}  \cdot  3 \left( H+  \frac{1}{\lambda_{\min} }\right)  \cdot \sqrt{SA K} \cdot   \frac{ (\lambda_{\max} H +1)  \ln K}{ \ln(\ln(K) +1)} ,
\end{align*}
where the first inequality follows from $L(\delta) >1$, the second inequality is due to \eqref{eq:cond-exp}, and the last inequality is due to Proposition~\ref{prop:proberror}. The proof of Proposition~\ref{prop:rateerror} is complete. } \qed


\subsection{Proof of Theorem~\ref{thm:lowerbound}} \label{sec:proof-LB}

We adapt the proof of Theorem 9 in \cite{domingues2021episodic} for episodic RL in discrete-time MDPs. The main thrust of their proof is to design a class of hard MDP instances that are based on the construction of MDPs in \cite[Chapter 38.7]{lattimore2020bandit} and \cite{Jaksch2010}.
We make necessary modifications of such MDP instances to fit into the continuous-time setting and prove our regret lower bound.

\noindent\paragraph{Proof of Theorem~\ref{thm:lowerbound}.}
Because the proof is similar to \cite{domingues2021episodic}, we only provide the key steps.
\begin{itemize}
\item \textit{Step 1. Constructing CTMDP instances}.

Given $S$ and $A$, by Assumption~\ref{assume:LB}, we can use $S-2$ states to construct a full A-ary tree of depth $d-1$. In this rooted tree, each node has exactly $A$ children and the total number of nodes is given by $\sum_{i=0}^{d-1}A^i = S-2$. The remaining two states are denoted by $s_g$ (`good' state) and $s_b$ (`bad' state), both of which are absorbing states. The transitions are deterministic within the tree: the initial state of the CTMDPs is the root of the tree, and given a node/state, action $a$ leads to the $a-$th child. Let $L$ be the number of leaves, and the collection of the leaves are denoted by  $\mathcal{L} = \{s_1, \ldots, s_L\}.$ For each $s_i \in \mathcal{L}$,  any action will lead to a transition to either the good state or the bad state according to
\begin{align*}
p(s_g | s_i, a) = \frac{1}{2} + \epsilon(s_i,a ), \quad p(s_b | s_i, a)  = \frac{1}{2} - \epsilon(s_i, a),
\end{align*}
where the function $\epsilon$ will be specified later. The reward function $r(s,a)=1$ if $s=s_g$ and it is zero otherwise.
The holding time for any state--action pair is exponentially distributed with rate $\lambda_{\max}$, except when the state is $s =s_g$ or $s=s_b$. Let $\mathcal{M}_0$ be the CTMDP with $\epsilon(s,a)=0$ for all $(s,a) \in \mathcal{L} \times \mathcal{A}$. For the elements (state-action pairs) in the set $\mathcal{J} := \mathcal{L} \times \mathcal{A}$ we order them from 1 to $|\mathcal{J}|$. Let $\mathcal{M}_j$, $j\geq 1$, be the CTMDP with $\epsilon (s,a) = \Delta$ for the $j$-th state--action pair in  $\mathcal{J} $ and $\epsilon (s,a) = 0$ otherwise,  where $\Delta$ is some parameter to be tuned later.

\item \textit{Step 2. Computing the expected regret of an \textbf{algo} in $\mathcal{M}_j$.}

Given a learning algorithm \textbf{algo}, for notational simplicity, denote by $R_j$ its expected regret for the CTMDP $\mathcal{M}_j$. The optimal cumulative expected reward over $[0, H]$ in the CTMDP $\mathcal{M}_j$ for $j \ge 1$ is given by
\begin{align*}
\mathbb{E} [ (H - \gamma_{d})^+]   \cdot (1/2 + \Delta),
\end{align*}
where $\gamma_{d}$ is the sum of $d$ i.i.d exponential random variables and it follows the Erlang distribution with shape parameter $d$ and rate parameter $\lambda_{\max}.$ This is because the optimal policy for $\mathcal{M}_j$ is simply to traverse the tree to reach the leaf state (corresponding to the state in the $j$-th state--action pair in $\mathcal{J}$ for which $\epsilon$ equals to $\Delta$) in $d-1$ transitions, and then take the corresponding action to reach the good state $s_g$ in one more transition (with probability $1/2 + \Delta$) to earn the reward. Denote by $\mathbb{P}_j$ and $\mathbb{E}_j$, $j\geq1$,  the probability measure and expectation in the CTMDP $\mathcal{M}_j$ by following the \textbf{algo}, and by $\mathbb{P}_0$ and $\mathbb{E}_0$ the corresponding operators in $\mathcal{M}_0.$ Then
the cumulative expected reward of the \textbf{algo} in $\mathcal{M}_j$ in episode $k$  is given by
\begin{align*}
\mathbb{E} [ (H - \gamma_d)^+]   \cdot   \mathbb{E}_j[1_{x_{d}^k = s_g} ],
\end{align*}
where $x_{d}^k$ is the state after $d$ transitions in episode $k$.
It follows that the expected regret is given by
\begin{align*}
R_j & = K \mathbb{E} [ (H - \gamma_d)^+]   \cdot (1/2 + \Delta) -  \mathbb{E} [ (H - \gamma_d)^+]   \cdot \sum_{k=1}^K   \mathbb{E}_j[1_{x_{d}^k = s_g} ] \\
& = K \mathbb{E} [ (H - \gamma_d)^+]   \cdot (1/2 + \Delta) -  \mathbb{E} [ (H - \gamma_d)^+]   \cdot (K/2 +  \Delta  \mathbb{E}_j\mathbf{T}_j ) \\
& = K \mathbb{E} [ (H - \gamma_d)^+]   \Delta \cdot ( 1 -   \frac{1}{K}  \mathbb{E}_j \mathbf{T}_j),
\end{align*}
where $\mathbf{T}_j \le K$ denotes the total number of visits to the state--action pair $j \in \mathcal{J}$ during  all the $K$ episodes while the \textbf{algo} is applied to $\mathcal{M}_j$. Hence, noting that  $|\mathcal{J}| = LA$, we conclude  that the maximum regret of the \textbf{algo} over all possible $\mathcal{M}_j$ for $j \in \mathcal{J}$ is lower bounded by
\begin{align}\label{eq:Rj}
\max_{j \in \mathcal{J} } R_j \ge \frac{1}{LA} \sum_{j \in  \mathcal{J}} R_j  = K \mathbb{E} [ (H - \gamma_d)^+]   \Delta \cdot \left( 1 -   \frac{1}{KLA}  \sum_{j \in \mathcal{J}} \mathbb{E}_j \mathbf{T}_j \right).
 \end{align}

 \item \textit{Step 3. Upper bounding $ \sum_{j \in \mathcal{J}} \mathbb{E}_{j} \mathbf{T}_j$.}

 Fix $j \in \mathcal{J}$. Since $\mathbf{T}_j/K \in [0,1 ]$, one can show that
 $kl( \frac{1}{K} \mathbb{E}_0[\mathbf{T}_j], \frac{1}{K} \mathbb{E}_j[\mathbf{T}_j]) \le \text{KL} (\mathbb{P}_0, \mathbb{P}_j)$,
 where \text{KL} denotes the Kullback--Leibler divergence between two probability measures and $kl (p, q)$ denotes the KL divergence between two Bernoulli distributions with success probabilities $p$ and $q$ respectively \citep{garivier2019explore}.
It then follows from Pinsker's inequality that
 \begin{align*}
  \frac{1}{K} \mathbb{E}_j[\mathbf{T}_j] \le  \frac{1}{K} \mathbb{E}_0 [\mathbf{T}_j] + \sqrt{\frac{1}{2} \text{KL} (\mathbb{P}_0, \mathbb{P}_j)}.
 \end{align*}
Because $\mathcal{M}_0$ and $\mathcal{M}_j$ only differ when the state--action pair $j$ is visited and the difference is only in the transition probabilities to the good or bad states, one can prove that
 \begin{align*}
  \text{KL} (\mathbb{P}_0, \mathbb{P}_j) = \mathbb{E}_0 [\mathbf{T}_j] \cdot kl(1/2, 1/2 + \Delta) \le \mathbb{E}_0 [\mathbf{T}_j] \cdot  4 \Delta^2,
 \end{align*}
when $\Delta \le 1/4.$ Hence,
 \begin{align*}
  \frac{1}{K} \mathbb{E}_j[\mathbf{T}_j] \le  \frac{1}{K} \mathbb{E}_0 [\mathbf{T}_j] + \sqrt{2} \Delta \sqrt{\mathbb{E}_0 [\mathbf{T}_j]}.
 \end{align*}
Note that $\sum_{j \in \mathcal{J}} \mathbf{T}_j \le K$. It follows from the Cauchy--Schwarz inequality that
 \begin{align} \label{eq:sumTj}
  \frac{1}{K} \sum_{j \in \mathcal{J}} \mathbb{E}_j[\mathbf{T}_j] \le  1 + \sqrt{2} \Delta \sum_{j \in \mathcal{J}}  \sqrt{\mathbb{E}_0 [\mathbf{T}_j]} \le 1+  \sqrt{2} \Delta \sqrt{LAK}.
 \end{align}

  \item \textit{Step 4. Optimizing $\Delta$ and finishing the proof.}

We  infer from \eqref{eq:Rj} and \eqref{eq:sumTj} that
\begin{align*}
\max_{j \in \mathcal{J}} R_j \ge K \mathbb{E} [ (H - \gamma_d)^+]   \Delta \cdot \left( 1 -   \frac{1}{LA} (1 +  \sqrt{2} \Delta \sqrt{LAK}) \right).
 \end{align*}
This lower bound is maximized by taking $\Delta = \frac{1}{2\sqrt{2}} (1- \frac{1}{LA}) \sqrt{\frac{LA}{K}} \le 1/4$ (by $K \ge SA/2$). Hence
\begin{align*}
\max_{j \in \mathcal{J}} R_j \ge \frac{1}{4 \sqrt{2}} \mathbb{E} [ (H - \gamma_d)^+]   \cdot \left( 1 -   \frac{1}{LA} \right)  \sqrt{LAK} .
 \end{align*}
Because $A \ge 2$ and $S \ge 6$, we have $L = \frac{1}{A} + (1- \frac{1}{A}) (S-2) \ge S/4$. So we obtain
 \begin{align*}
\max_{j \in \mathcal{J}} R_j \ge \frac{1}{12 \sqrt{2}} \mathbb{E} [ (H - \gamma_d)^+]   \cdot \sqrt{SAK} .
 \end{align*}
 Therefore, there exists a CTMDP $\mathcal{M}_j$ for some $j \in \mathcal{J}$ such that \eqref{eq:regret-lowerbound} holds. Finally, \eqref{eq:regret-lowerbound2} follows  from  \eqref{eq:regret-lowerbound} by Jensen's inequality. The proof is complete. \qed
 \end{itemize}



\section{Conclusions}\label{sec:conclusion}
In this paper we study RL for tabular CTMDPs with unknown parameters in the finite-horizon, episodic setting. We develop a learning algorithm and establish a worst-case regret upper bound. Meanwhile, we prove a regret lower bound,  showing that the square-root regret rate achieved by our proposed algorithm actually  has the optimal dependance on the numbers of episodes and actions. Numerical experiments are conducted to illustrate the performance of our learning algorithm.

Our work serves as a first step towards a better understanding of  efficient episodic RL for CTMDPs. It is an open question how to improve our worst-case (or instance-independent) regret bounds and narrow the gap between the upper and lower bounds. There are also many other open problems, including, to name but a few, instance-dependent regret bounds, learning in CTMDPs with large or infinite state spaces, and episodic RL for semi-Markov decision processes that allow  general holding time distributions. We leave them for future research.


\bibliographystyle{chicago}
\bibliography{RL2}

\end{document}